\documentclass[journal]{IEEEtran}

\usepackage{cite}
\usepackage{amssymb}
\usepackage{amsmath}
\usepackage{mathrsfs}
\usepackage{algorithm}
\usepackage{subfigure}
\usepackage{graphicx}
\usepackage[breaklinks=true,letterpaper=true,colorlinks,bookmarks=false,citecolor=blue]{hyperref}
\usepackage{booktabs}
\usepackage{multirow}
\usepackage{multicol}
\usepackage{float}
\usepackage{color}
\usepackage{array}
\usepackage{colortbl}
\usepackage[keeplastbox]{flushend}

\usepackage{threeparttable}  

\newcounter{magicrownumbers}  
\newcommand\rownumber{\stepcounter{magicrownumbers}\arabic{magicrownumbers}}

\begin{document}

\title{Mini-Unmanned Aerial Vehicle-Based Remote Sensing: Techniques, Applications, and Prospects}

\author{Tian-Zhu Xiang,
	Gui-Song Xia,~\IEEEmembership{Senior Member,~IEEE,}
	and~Liangpei Zhang,~\IEEEmembership{Fellow,~IEEE}
	\thanks{Tian-Zhu Xiang, Gui-Song Xia and Liangpei Zhang are with the State Key Lab. \emph{LIESMARS}, Wuhan University, Wuhan, 430079, China. E-mail: \{\emph{tzxiang, guisong.xia, zlp62}\}@whu.edu.cn. This work was supported in part by the National Natural Science Foundation of China under Grants 61771350, 61871299 and 41820104006, in part by the Outstanding Youth Project of Hubei Province under Contract 2017CFA037. (\emph{Corresponding author: Gui-Song Xia.})}
	\thanks{Manuscript received January **, 2019; revised ** **, 2019.}
}

\markboth{Journal of \LaTeX\ Class Files,~Vol.~14, No.~8, ***~2019}%
{Shell \MakeLowercase{\textit{et al.}}: Bare Demo of IEEEtran.cls for IEEE Journals}

\maketitle

\begin{abstract}
	\textcolor{blue}{This is the preprint version, to read the final version please go to IEEE Geoscience and Remote Sensing Magazine on IEEE Xplore.} The past few decades have witnessed the great progress of unmanned aircraft vehicles (UAVs) in civilian fields, especially in photogrammetry and remote sensing. In contrast with the platforms of manned aircraft and satellite, the UAV platform holds many promising characteristics: flexibility, efficiency, high-spatial/temporal resolution, low cost, easy operation, etc., which make it an effective complement to other remote-sensing platforms and a cost-effective means for remote sensing. In light of the popularity and expansion of UAV-based remote sensing in recent years, this paper provides a systematic survey on the recent advances and future prospectives of UAVs in the remote-sensing community. Specifically, the main challenges and key technologies of remote-sensing data processing based on UAVs are discussed and summarized firstly. Then, we provide an overview of the widespread applications of UAVs in remote sensing. Finally, some prospects for future work are discussed. We hope this paper will provide remote-sensing researchers an overall picture of recent UAV-based remote sensing developments and help guide the further research on this topic.
\end{abstract}


%
\IEEEpeerreviewmaketitle

\section{Introduction}
\label{sec:intro}

\IEEEPARstart{I}{n} recent years, with the rapid development of economy and society, great changes have been taken place on the earth's surface constantly. Thus, for the remote-sensing community, it is in great demand to acquire remote-sensing data of interesting regions and update their geospatial information flexibly and quickly~\cite{PerryRyan2011,ZhangJohn2012,LiuChen2014}.

\subsection{Motivation and objective}

\begin{figure}[t]
	\centering
	\includegraphics[width=0.98\linewidth]{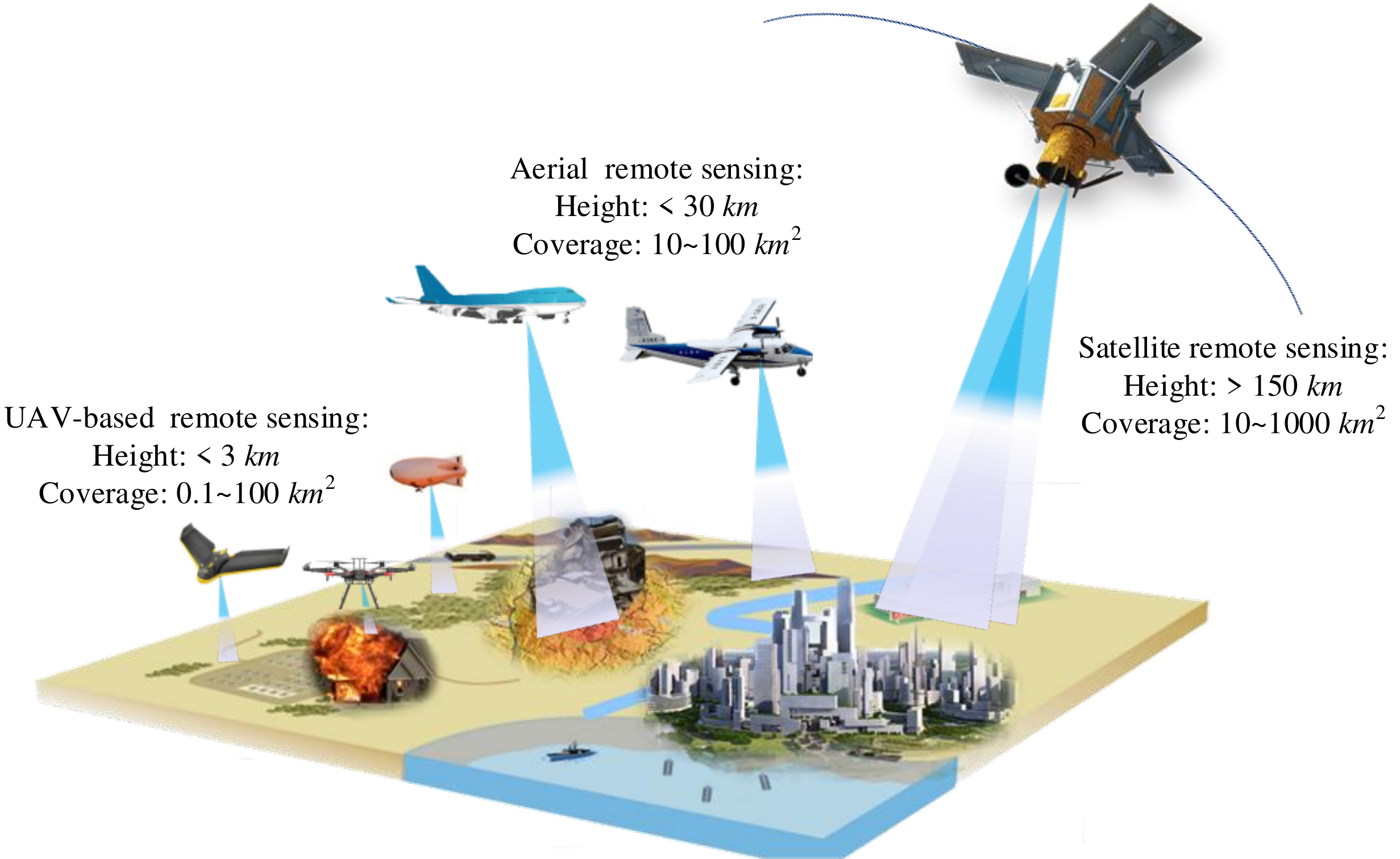}
	\caption{Remote sensing platforms of satellite, manned aviation and low-altitude UAV.}
	\label{fig:threeplatforms}
\end{figure}

The main ways of earth observation and geospatial information acquisition are satellite (shown in Tab.~\ref{tab:SatRS}), manned aviation and low-altitude remote sensing~\cite{TOTHRS2016}, shown in Fig.~\ref{fig:threeplatforms}. Remote sensing based on satellite and manned aircraft often have the advantages of large-area or regional remote sensing emergency monitoring with multi-sensors~\cite{JonHRRS2012}. However, due to the satellite orbit, airspace of take-off and landing, meteorological conditions, etc., these two ways have some limitations, discussed as follows.

\subsubsection{\textbf{Timeliness of data}}
In many time-critical remote-sensing applications, it is of great importance to timely acquisition of remote sensing data with high temporal resolution. For instance, in emergency remote sensing, \emph{e.g.} earthquake, flood and landslide, fast response is the prerequisite~\cite{GomezPurdie2016}. It is necessary to collect remote sensing data of disaster area promptly and frequently for dynamical monitoring and analysis of disaster situation. In addition, precision agriculture requires short revisit times to examine within-field variations of crop condition, so as to respond to fertilizer, pesticide, and water needs~\cite{HuntRondon2017}.

However, although the revisit cycles of satellite sensors have significantly decreased to one day, shown in Tab.~\ref{tab:SatRS}, due to the launch of satellite constellations and the increasing number of operating systems, it may not be easy to provide response of abrupt changes quickly and multiple acquisition per day.
The manned aviation platforms are capable of collecting high-resolution data without the limitation of revisit periods, while they suffer from low maneuverability, high launch/flight costs, limitation of airspace and complex logistics. Besides, the data from these two platforms is often severely limited by weather conditions (\emph{e.g.} cloud cover, haze and rain), which affects the availability of remote-sensing data~\cite{DavidPA2013}.

\subsubsection{\textbf{Spatial resolution}}
\begin{figure}[tp]
	\centering
	\subfigure[Dam/Bridge crack detection]{
		\includegraphics[height=0.3\linewidth, width=0.4\linewidth]{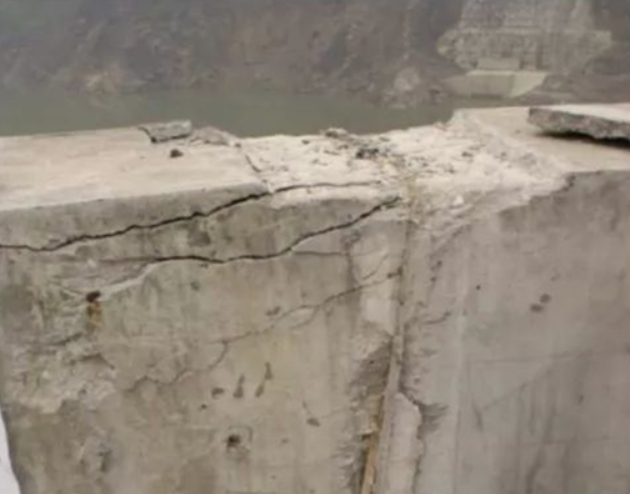}}
	\subfigure[Buddha reconstruction]{
		\includegraphics[height=0.3\linewidth, width=0.4\linewidth]{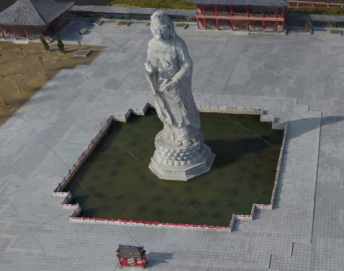}} \\
	\subfigure[Pine nematode detection]{
		\includegraphics[height=0.3\linewidth, width=0.4\linewidth]{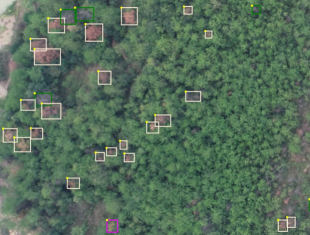}}
	\subfigure[Counting of crop plants]{
		\includegraphics[height=0.3\linewidth, width=0.4\linewidth]{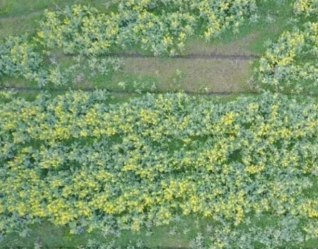}}
	\caption{Examples of ultra-high spatial resolution remote sensing.}
	\label{fig:spatialReso}
\end{figure}

Remote sensing data with ultra-high spatial resolution (\emph{e.g.} centimeter-level) plays significant roles in some fine-scale remote sensing applications, such as railway monitoring, dam/bridge crack detection, reconstruction and restoration of cultural heritage~\cite{LouisetPamart2016}. Besides, numerous studies have reported that images with centimeter-level spatial resolution (up to 5 \emph{cm} or more) hold the potential for studying spatio-temporal dynamics of individual organisms~\cite{MateseTD2015}, mapping fine-scale vegetation species and their spatial patterns~\cite{LaliberteRango2009}, estimating landscape metrics for ecosystem~\cite{AndersonGaston2013}, monitoring small changes of coasts by erosion~\cite{PrahaladIs2015}, etc. Some examples are shown in Fig.~\ref{fig:spatialReso}.

Currently, satellite remote sensing can provide high-spatial-resolution images of up to 0.3 \emph{m}, but it remains not to meet the requirements of aforementioned applications. Manned aviation remote sensing is capable of collecting ultra-high spatial resolution data, while it is restricted by operational complexity, costs, flexibility, safety and cloud cover.

\begin{table}[!tb]
	\renewcommand\arraystretch{1.5}
	\caption{Some examples of optical satellite remote sensing.}
	\label{tab:SatRS}
	\centering
	\begin{threeparttable}
	\begin{tabular}{c m{1.7cm}<{\centering} m{1.9cm}<{\centering} m{1.3cm}<{\centering}}
		\hline
		Name                  & GSD of PAN/MS (m)     & Temporal resolution (day)     & Nations\\
		\hline
		Planet Labs             & 0.72$\sim$5/-             & 1        & USA    \\
		GF-2                       & 0.8/3.2                & 5        & China  \\   
		Surperview-1          & 0.5/2                  & 4        & China  \\       
		Worldview-4          & 0.31/1.24            & 1-3     & USA    \\     
		Geoeye-1                & 0.41/1.65            & 2-3      & USA    \\     
		Pleiades                  & 0.5/2                   & 1         & France \\ 
		SPOT-7                   & 1.5/6                   & 1        & France \\ 
		KOMPSAT-3A       & 0.4/1.6                & 1         & Korean \\ 
		\hline
	\end{tabular}
\begin{tablenotes}
	\footnotesize
	\item[*] GSD: Ground sample distance; PAN: Panchromatic image; MS: Multi-spectral image.
\end{tablenotes}
\end{threeparttable}
\end{table}

\subsubsection{\textbf{Data quality and information content}}
Remote sensing data from satellite and manned aircraft platforms are susceptible to cloud conditions and atmosphere, which attenuate electromagnetic waves and cause information loss and data degradation. While low-altitude platforms have the advantage of flying closer to the ground object, which mitigate the effects of cloud and atmosphere significantly. Therefore, low-altitude remote sensing has the advantage of collecting high quality data with rich information and high definition, which benefits for image interpretation. Meanwhile, there is no need for atmospheric corrections as it would be in traditional platforms~\cite{TelmoHSIUAV2017}.
	
Besides, satellite and manned aircraft platforms mainly focus on high-resolution orthophotos, and they are unable to provide high-resolution multi-view facade and occlusion area images, which play a central role in three-dimension (3D) fine modeling~\cite{NexRemondino2014}. Moreover, it has been demonstrated that multi-view information of ground object is beneficial for analyzing the anisotropic characteristics of its reflectance and further improving remote sensing image classification~\cite{LiuFCNMV2018}.

\subsubsection{\textbf{Small area remote sensing}}
Satellite and manned aircraft platforms often run on fixed orbits or operate along the preset regular paths. However, in many small-area remote sensing applications, \emph{e.g.}, small town planning, tiny islands mapping, urban small-area geographic information update, archeology, agricultural breeding and infrastructure damage detection, there is a demand that collecting data along the irregular planning routes, or modifying route temporarily and taking hover observation according to tasks. The lack of flexibility makes utilization of traditional platforms challenging. The safety of pilots and cost also limit the adoption of manned aircraft platforms. In addition, traditional platforms may be difficult to acquire data in dangerous, difficult-to-access or harsh environments, such as polar remote sensing~\cite{Leary2017}, monitoring of nuclear radiation, volcanoes and toxic liquids~\cite{GomezPurdie2016}.

\vspace{6pt}

Consequently, to compensate these deficiencies, remote-sensing scientists propose some low-altitude remote sensing platforms, such as light aircraft platforms~\cite{Graham1988}, remote-control aircrafts or kites~\cite{Verhoeven2009}, and unmanned aerial vehicles (UAVs)~\cite{Colomina2014}. Due to the unique advantages, \emph{e.g.} flexibility, maneuverability, economy, safety, high-spatial resolution and data acquisition on demand, UAVs have been recognized as an effective complement to traditional platforms. In recent years, the boom of UAV technology and the advance of small-sized, low-weight and high-detection-precision sensors equipped on these platforms make the UAV-based remote sensing (UAV-RS) a very popular and increasingly used remote-sensing technology.

It is also worth noting that the continuous advance of satellite constellations will improve the spatial/temporal resolution and data acquisition cost of satellite remote sensing. Therefore, in the future, it can be predicted that UAVs will replace manned aircraft platforms and become the main means for remote sensing together with satellite platforms~\cite{LiaoZhou2016}.

Considering the rapid evolution of UAV-RS, it is essential to take a comprehensive survey on the current status of UAV-based remote sensing, in order to gain a clearer panorama in UAV-RS and promote further progress. Thus, this paper presents a specific review of recent advances on technologies and applications from the past few years. Some prospects for future research are also addressed.

\begin{figure}[!htb]
	\begin{center}
		\includegraphics[width=0.9\linewidth]{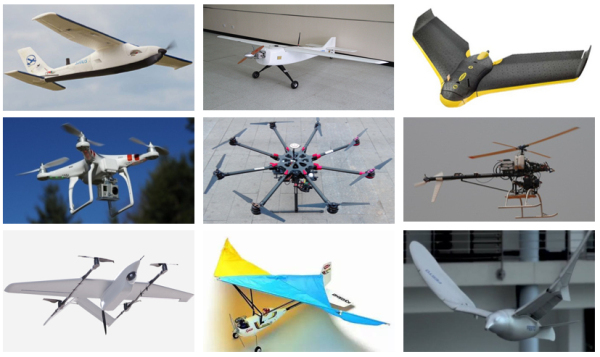}
	\end{center}
	\caption{Some examples of mini-UAVs for remote sensing. Top: fixed-wing UAVs. Middle: rotary-wing UAVs and unmanned helicopters. Bottom: Hybrid UAVs, umbrella-UAVs, and bionic-UAVs.}
	\label{fig:UAVsExample}
\end{figure}

\begin{figure}[tb]
	\begin{center}
		\includegraphics[width=0.9\linewidth]{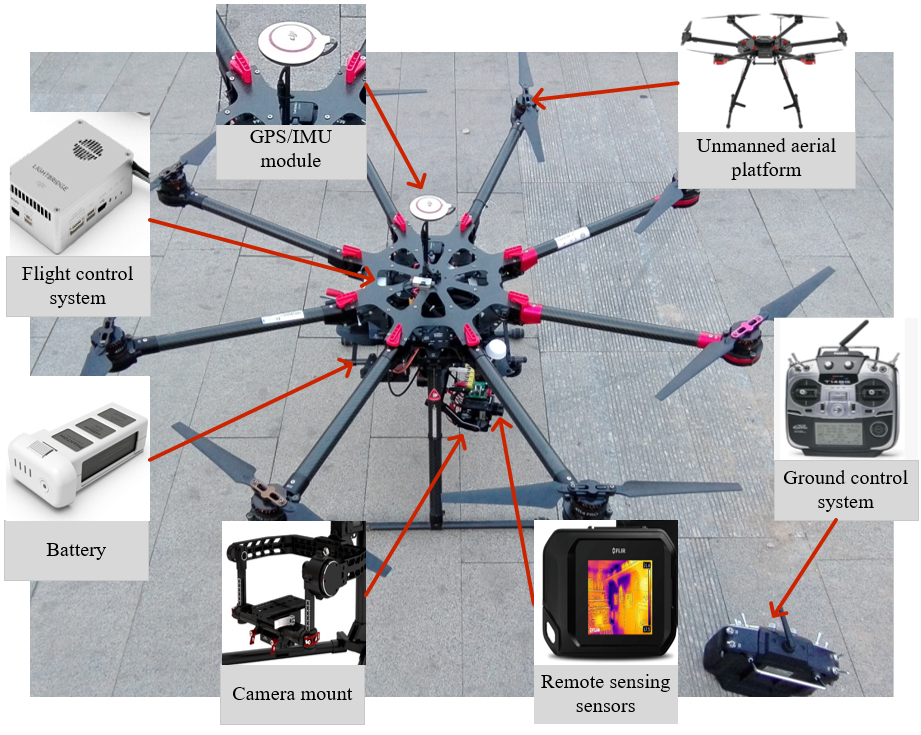}
	\end{center}
	\caption{An example of the rotary-wing UAV-based remote sensing data acquisition platform.}
	\label{fig:RotaryUAVRSExample}
\end{figure}

In this paper, we focus on the Mini-UAV which features less than thirty kilograms of maximum take-off weight~\cite{AndersonGaston2013, Colomina2014}, since this type of UAVs, more affordable, easier to carry and use than large-sized UAVs, is one of the most widely used types in remote-sensing community. Some examples of mini-UAVs is shown in Fig.~\ref{fig:UAVsExample}. A simple example of rotary-wing UAV-RS system is shown in Fig.~\ref{fig:RotaryUAVRSExample}. In this system, the infrared camera is equipped on an eight-rotor UAV to acquire thermal radiation data around heat-supply pipelines for detection of heat leakage. Recognizing space limitations, more detailed description of the unmanned aircraft and remote-sensing sensors specially designed for UAV platforms can be found in~\cite{Colomina2014, Pajares2015}.

\subsection{Related to previous surveys}

\begin{table*}[!tb]
	\scriptsize  
	\renewcommand\arraystretch{1.5}
	\caption{Summarization of a number of related surveys on UAV-RS in recent years$^{*}$.}
	\label{tab:surveyset}
	\centering
	\begin{threeparttable}
	\begin{tabular}{ m{0.18cm}<{\centering} | m{6.5cm}<{\centering} | m{0.4cm}<{\centering} | m{0.4cm}<{\centering} | m{0.8cm}<{\centering} | m{7cm}<{\centering}}  
		\toprule
		No.        & Survey Title         & Ref.       & Year      & Published     & Content	\\
		\midrule
		
		
		\rownumber  & Overview and Current Status of Remote Sensing Applications Based on Unmanned Aerial Vehicles (UAVs)
		& ~\cite{Pajares2015}           & 2015  & PERS
		& A broad review of current status of remote sensing applications based on UAVs \\ \cline{1-6} 
		
	    \rownumber  & Unmanned Aerial Systems for Photogrammetry and Remote Sensing: A Review  
		& ~\cite{Colomina2014}          & 2014  & ISPRS JPRS  
		& A survey of recent advances in UAS and its applications in Photogrammetry and Remote Sensing \\ \hline
		
		\rownumber  & Hyperspectral Imaging: A Review on UAV-Based Sensors, Data Processing and Applications for Agriculture and Forestry
		& ~\cite{TelmoHSIUAV2017}       & 2017  & RS
		& A survey of UAV-based hyperspectral remote sensing for agriculture and forestry \\ \cline{1-6}
		
		\rownumber  & UAS, Sensors, and Data Processing in Agroforestry: A Review Towards Practical Applications
		& ~\cite{LuisUAVAG2017}         & 2017  & IJRS  
		& A survey of data processing and applications of UAS and sensors in agroforestry, and some recommendations towards UAS platform selection \\ \cline{1-6}
	
		\rownumber  & Forestry Applications of UAVs in Europe: A Review
		& ~\cite{TorresanBerton2017}    & 2017  & IJRS
		& An overview of applications of UAVs in forest research in Europe, and an introduction of the regulatory framework for the operation of UAVs in the European Union  \\ \cline{1-6}	
			
		\rownumber  & UAVs as Remote Sensing Platform in Glaciology: Present Applications and Future Prospects
		& ~\cite{BhardwajSam2016}       & 2016  & RSE 
		& A survey of applications of UAV-RS in glaciological studies, mainly in polar and alpine applications \\ \cline{1-6}		

		\rownumber  & Recent Applications of Unmanned Aerial Imagery in Natural Resource Management
		& ~\cite{ShahbaziTM2014}           & 2014  & GISRS 
		& A comprehensive review of applications of unmanned aerial imagery for the management of natural resources.\\ \cline{1-6}		 

        \rownumber  & Small-scale Unmanned Aerial Vehicles in Environmental Remote Sensing: Challenges and Opportunities
        & ~\cite{PerryRyan2011}            & 2011  & GISRS 
        & An introduction of challenges involved in using small UAVs for environmental remote sensing \\ \cline{1-6}     
 
 
        \rownumber  & Recent Developments in Large-scale Tie-point Matching
        & ~\cite{Hartmann2016}             & 2016   & ISPRS JPRS   
        & A survey of large-scale tie-point matching in unordered image collections \\ \cline{1-6}
        
        \rownumber  & State of the Art in High Density Image Matching
        & ~\cite{RemondinoS2014}           & 2014   & PHOR  
        & A review and comparative analysis of four dense image-matching algorithms, including SURE (semi-global matching), MicMac, PMVS and Photoscan \\ \cline{1-6}
        
        \rownumber  & Development and Status of Image Matching in Photogrammetry
        & ~\cite{Gruen2012}                & 2012   & PHOR  
        & A comprehensive survey of image matching techniques in photogrammetry over the past 50 years \\ \hline
    
            
        \rownumber  & Review of the Current State of UAV Regulations
        & ~\cite{StockerRegu2017}          & 2017   & RS
        & A comprehensive survey of civil UAV regulations on the global scale from the perspectives of past, present, and future development \\ \cline{1-6}
       
        \rownumber  & UAVs:Regulations and Law Enforcement 
        & ~\cite{ArthurRegula2017}         & 2017   & IJRS
        & An introduction to the development of legislations of different countries regarding UAVs and their use \\ \cline{1-6}
         
        \rownumber  & Unmanned Aerial Vehicles and Spatial Thinking: Boarding Education With Geotechnology and Drones
        & ~\cite{FombuenaUAVEDU2017}        & 2017   & GRSM
        & A review of current status of geosciences and RS education involving UAVs \\ \cline{1-6}
        
        \rownumber  & Unmanned Aircraft Systems in Remote Sensing and Scientific Research: Classification and Considerations of Use
        & ~\cite{WattsAmbrosia2012}        & 2012   & RS
        & An introduction to UAS platform types, characteristics, some application examples and current regulations \\ \cline{1-6}

        \rownumber  & \textbf{Mini-UAV-based Remote Sensing: Techniques, Applications and Prospectives}
        & -        & \textbf{2019}   & \textbf{Ours}
        & \textbf{A comprehensive survey of mini-UAV-based remote sensing, focusing on techniques, applications and future development} \\
            
	\bottomrule	  
	\end{tabular}
    \begin{tablenotes}
	\scriptsize 
	\item[*] This table only shows surveys published in top remote-sensing journals.
	\item[*] PERS: Photogrammetric Engineering and Remote Sensing; ISPRS JPRS: ISPRS Journal of Photogrammetry and Remote Sensing; RS: Remote Sensing; IJRS: International Journal of Remote Sensing; RSE: Remote Sensing of Environment; GISRS: GIScience \& Remote Sensing; PHOR: The Photogrammetric Record. GRSM: IEEE Geoscience and Remote Sensing Magazine.
    \end{tablenotes}
    \end{threeparttable}
\end{table*}

A number of representative surveys concerning UAV-based remote sensing have been published in the literature, as summarized in Tab.~\ref{tab:surveyset}.
    	
These include some excellent surveys on the hardware development of unmanned aerial systems, \emph{e.g.} unmanned aircrafts and sensors~\cite{WattsAmbrosia2012, Colomina2014, Pajares2015, TelmoHSIUAV2017, LuisUAVAG2017}, less attention has been paid to the advance of UAV data processing techniques.	Some surveys focus on specific aerial remote-sensing data processing, such as image matching~\cite{Hartmann2016, Gruen2012} and dense image matching~\cite{RemondinoS2014}, which are not specifically for UAV data processing. Although the research reviewed in~\cite{Colomina2014, LuisUAVAG2017} presents some UAV data processing technologies, \emph{e.g.} 3D reconstruction and geometric correction, there still lack a complete survey of UAV data processing and its recent advances. In addition, recent striking success and potential of deep learning and related methods on UAV data geometric processing has not been well investigated.
		
Some surveys review specific applications of UAVs in remote-sensing community, such as agriculture~\cite{TelmoHSIUAV2017}, forestry~\cite{LuisUAVAG2017,TorresanBerton2017}, natural resource management~\cite{ShahbaziTM2014}, environment~\cite{PerryRyan2011} and glaciology~\cite{BhardwajSam2016}. Besides, \cite{Colomina2014} and \cite{Pajares2015} provides a comprehensive reviews of applications of UAV-RS, which also include the advance of remote-sensing sensors and regulations. However, recent developments in the technology of UAV-RS have opened up some new possibilities of applications, \emph{e.g.} pedestrian behavior understanding~\cite{Robicquet2016}, intelligent driving and path planning~\cite{WallarDriving2018}, which have not been reviewed.

\subsection{Contributions}

Considering the problems discussed above, it is imperative to provide a comprehensive surveys of UAV-RS, centering on UAV data processing technologies, recent applications and future directions, the focus of this survey. A thorough review and summarization of existing work is essential for further progress in UAV-RS, particularly for researchers wishing to enter the field. Extensive work on other issues, such as regulations~\cite{Colomina2014, StockerRegu2017, ArthurRegula2017}, operational considerations~\cite{WattsAmbrosia2012, AndersonGaston2013, LuisUAVAG2017}, which have been well reviewed in the literature, are not included.

Therefore, this paper is devoted to present: 
\begin{itemize}
	\item[-] A systematic survey of data processing technologies, categorized into eight different themes. In each section, we provide a critical overview of the state-of-the-art, illustrations, current challenges and possible future works;
	\item[-] A detailed overview of recent potential applications of UAVs in remote sensing;
	\item[-] A discussion of the future directions and challenges of UAV-RS from the point of view of platform and technology.
\end{itemize}

The remainder of this paper is organized as follows. The main challenges and technologies of UAV-RS data processing are reviewed and discussed in Section~\ref{sec:techniques}. The potential applications of UAVs in remote-sensing community are provided in Section~\ref{sec:applications}. In Section~\ref{sec:prospect}, the current problems and future development trend of UAV-RS are explored. At last, we conclude the paper in Section~\ref{sec:conclusions}.

\section{Techniques for data processing}
\label{sec:techniques}

In this section, main challenges of UAV data processing are briefly introduced. Then, we discuss the general processing framework and the key technologies as well as the recent improvements and breakthroughs of them.

\subsection{Main challenges} 
\label{Sec:Challenges}

Compared with satellite and manned aerial remote sensing, UAV-based remote sensing has incomparable advantages to provide a low-cost solution to collect data at spatial, spectral and temporal scales. However, it also faces some special challenges, due to the big differences with satellite and manned aerial remote sensing in platforms, flight height, sensors and photographic attitude, as well as external effects (\emph{e.g.} airflow).

\begin{itemize}
	\item[1)] \emph{\textbf{Non-metric camera problem.}} 
	Due to payload weight limitations, UAV-RS often adopts low-weight, small-size and non-metric (consumer-grade) cameras, which inevitably result in some problems.
	
	\begin{itemize}
		\item[-] Camera geometry issue. Camera factory parameters are generally inaccurate and often affected by extraneous factors (\emph{e.g.} camera shake). In addition, there are serious lens distortion in consumer-grade cameras, such as radial and tangential distortions. These problems reduce accuracy of data processing, especially in spatial resection and object reconstruction~\cite{SteveArko2015}. Thus, it is necessary to calibrate cameras strictly before data processing.
		
		\item[-] Rolling-shutter issue. Most UAVs are equipped with low-cost rolling-shutter cameras. Unlike global shutter, in rolling-shutter acquisition mode, each row is exposed in turn and thus with a different pose when the unmanned aircraft flies~\cite{KlingnerMFSFM2013}. In addition, moving rolling-shutter cameras often produce image distortions~\cite{DengUAVPA2018} (\emph{e.g.} twisting and slanting). These are beyond the conventional geometric models in 3D vision. Thus, new methods for rolling-shutter cameras are strongly desired.
		
		\item[-] Other issues, including noise, vignetting, blurring and color unbalancing which degrade image quality. 
    \end{itemize}

	\item[2)] \emph{\textbf{Platform instability and vibration effects.}} 
	The weak wind resistance make the light-weight, small-size UAVs collect remote-sensing data with unstable sensor positions, which affects data quality~\cite{ZhangJohn2012, AndersonGaston2013}. 
	
	\begin{itemize}
		\item[-] The data is often acquired with irregular air lines, even curved lines. It results in image overlap inconsistency, which may causes failure image connection in aerial triangulation, especially between flight strips. Meanwhile, it also leads to complex and unordered image correspondence, which makes it difficult to determine which pairs of images can be matched.
		
		\item[-] Variable attitudes of sensors may result in large rotation and tilt variations among images, and thus bring about obvious image affine deformation. In addition, it can also result in large non-uniformity of scale and illumination. These issues will be aggravated by complex terrain relief, and present challenges for image matching~\cite{JiangJiang2017}.
	\end{itemize}

	\begin{figure}[tb]
		\begin{center}
			\includegraphics[width=0.98\linewidth]{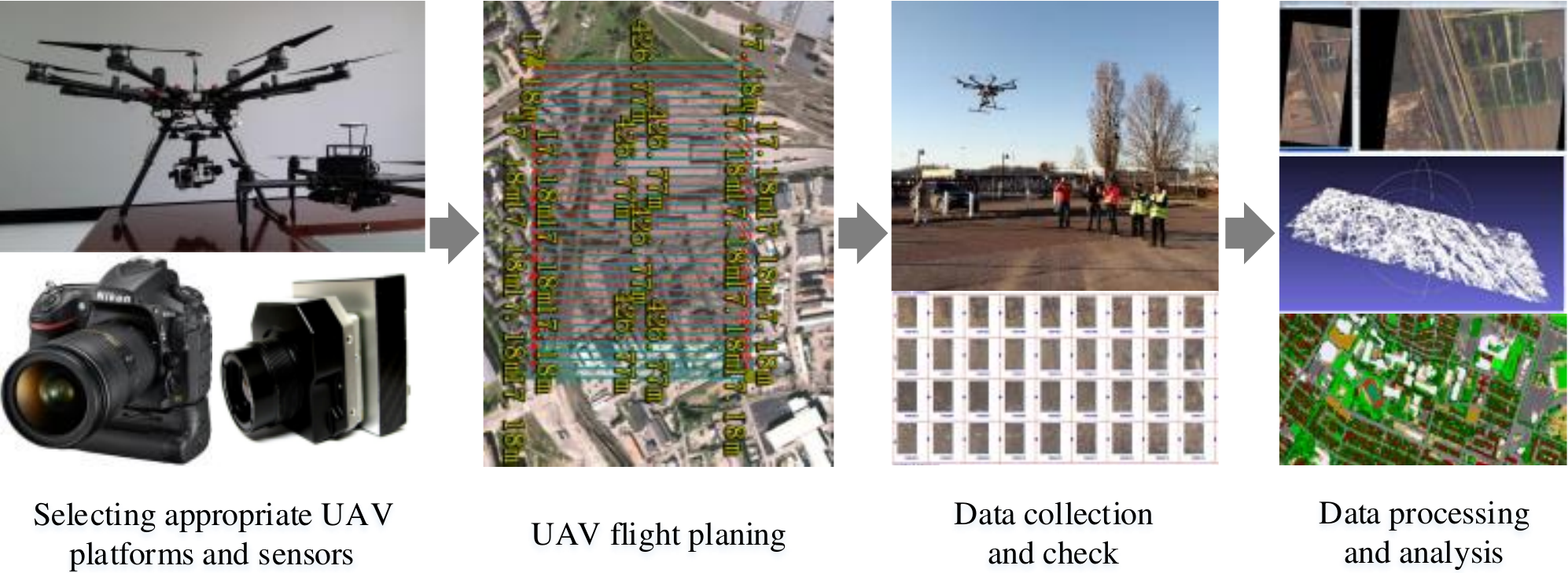}
		\end{center}
		\caption{General workflow of UAV-based remote sensing.}
		\label{fig:Workflow}
	\end{figure}

	\begin{figure*}[!htp]
		\begin{center}
			\includegraphics[width=0.98\linewidth]{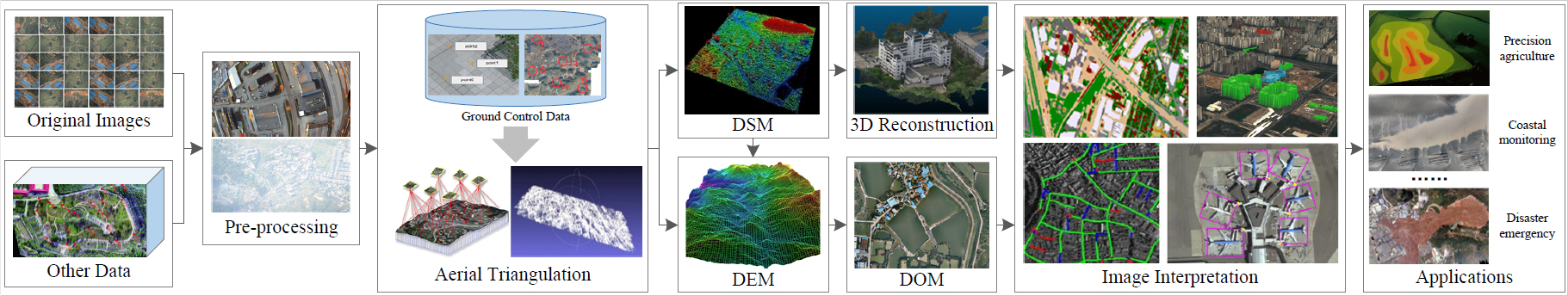}
		\end{center}
		\caption{General workflow of UAV-RS data processing.}
		\vspace{-0.3cm}
		\label{fig:Dataprocess}
	\end{figure*}

	\item[3)] \emph{\textbf{Large amount of images and high overlap.}}	
	The small field of views (FOVs) of cameras equipped on UAVs, along with low acquisition height, make UAVs need to capture more photographs than conventional platforms to ensure overlaps and coverage. Therefore, on one hand, it is common that some images only cover homogeneous areas with low textures, resulting in difficulties for feature detection. On the other hand, the large amount of images may result in large-scale tie-points, which increases the difficulty and time for image matching and aerial triangulation. Besides, to ensure overlaps, images are often acquired with high amounts of overlap, which may lead to short baselines and small base-height ratio. Thus, it may cause unstable aerial triangulation and low elevation accuracy.

	\item[4)] \emph{\textbf{Relief displacement.}}
	Due to the low acquisition altitudes relative to the variation in topographic relief, UAV image processing is prone to the effects of relief displacement ~\cite{KenChris2014prog}.  Such displacement can generally be removed by orthorectification if the digital elevation/surface model represents the terrain correctly. It remains challenging to handle the scenes with trees or buildings, because of the large local displacement and occlusion areas with no data. Besides,  effects will be obvious when mosaicking images with different amounts and directions of relief displacement, such as sudden break, blurring and ghosting. 
	
\end{itemize}

Due to the challenges discussed above, there exists large difficulties for traditional photogrammetric processing approaches designed for well-calibrated metric cameras and regular photography. Hence, rigorous and innovative methodologies are required for UAV data processing and have become a center of attention for researchers worldwide.

\subsection{General framework}

A general workflow of UAV-based remote sensing is shown in Fig.~\ref{fig:Workflow}. To conduct data acquisition, suitable UAV platforms and remote-sensing sensors are first selected according to remote-sensing tasks. More importantly, all the hardware needs to be calibrated, including cameras and multi-sensor combination, so as to determine spatial relationship of different sensors and remove geometric distortions caused by cameras. Then mission planning is designed based on topography, weather and lighting conditions in the study areas. Flight parameters, such as flight path, flying altitude, image waypoints, flight speed, camera length and exposure time, need to be carefully designed to ensure data overlaps, full coverage and data quality. Afterwards, remote-sensing data are often collected autonomously based on flight planning, or by the flexible control of the ground pilot. Data is checked and a supplementary photograph is performed if necessary. After data acquisition, a series of methods are performed for data processing and analysis.

To illustrate the UAV-based remote sensing data processing, we takes optical cameras, one of the most widely applied sensors, as an example. The general workflow of data processing can be seen in Fig.~\ref{fig:Dataprocess}. Specifically, 

\begin{itemize}
	\item[1)] \emph{Data pre-processing.} Images collected from UAV platforms often require pre-processing to ensure their usefulness for further processing, including camera distortion correction, image color adjustment, noise elimination, vignetting and blur removal~\cite{LeiUAVSR2018}. 
	
	\item[2)] \emph{Aerial triangulation,} also called structure from motion (SfM) in computer vision. It aims to recover the camera pose (position and orientation) per image and 3D structures (\emph{i.e.} sparse point clouds) from image sequences, which can also provide a large number of control points of orientation for image measurement. Data from GPS and inertial measurement unit is often used to initialize the position and orientation of each image. In computer vision, camera poses can be estimated based on image matching. Besides, image matching can also be adopted to generate a large number of tie-points and build connection relationships among images. Bundle adjustment is  used to optimize the camera positions and orientations and derive scene 3D structures. To meet requirements of  high-accuracy measurement,  the use of ground control points (GCPs) may be necessary for improving georeferencing, while it is a time-consuming and labor-intensive work.
	
	\item[3)] \emph{Digital surface model (DSM) generation and 3D reconstruction.} The oriented images are used to derive dense point clouds (or DSM) by dense image matching. DSM provides a detailed representation of the terrain surface. Combining with surface reconstruction and texture mapping, a 3D model of scene can be well reconstructed.
	
	\item[4)] \emph{Digital elevation model (DEM) and orthophoto generation.} Digital elevation model can describe the surface topography without effects of raised objects, such as trees and buildings. It can be generated from either sparse or dense point clouds. The former is with lower precision while higher efficiency than the latter. After that, each image can be orthorectified to eliminate the geometric distortion, and then mosaicked into a seamless orthonormal mosaic at the desired resolution.
	
	\item[5)] \emph{Image interpretation and application.} Based on orthophotos and 3D models, image interpretation are performed to achieve scene understanding, including image/scene classification, object extraction and change detection. Furthermore, the interpretation results are applied for various applications, such as thematic mapping, precision agriculture and disaster monitoring. 

\end{itemize}

In fact, regardless of the platform from which remote-sensing data is acquired (satellite, airborne, UAV, etc.), its interpretation methods are similar~\cite{TelmoHSIUAV2017}. Therefore, photogrammetric processing is the prominent concern regarding UAV-RS. It is challenging issue for traditional processing approaches. Methods specially designed for UAV-RS data processing are proposed to overcome issues of UAV-RS. Next, the related key technologies are reviewed and summarized.

\subsection{Camera calibration}
\label{Sec:Calib}

Different from the traditional remote-sensing data processing, camera calibration is essential for UAV-based remote sensing, due to the adoption of light-weight and non-metric cameras that have not been designed for photogrammetric accuracy~\cite{AasenBurkart2015}. Camera calibration aims to estimate camera parameters to eliminate the impact of lens distortion on images and extract metric information from 2D images~\cite{WangWZ2008kruppa}. In aerial triangulation, camera parameters, including intrinsic parameters (principal-point position and focal length) and lens distortion coefficients (radial and tangential distortion coefficients), are often handled by pre-calibration or on-the-job calibration. The former calibrates cameras before bundle adjustment, and the latter combines camera calibration parameters as unknowns into bundle adjustment for joint optimization and estimation. The combination of two options is also adopted for high-accuracy data processing~\cite{StrechaCalib2008}. On-the-job calibration is often sensitive to camera network geometry (\emph{e.g.} nadir and oblique acquisition) and the distribution and accuracy of ground control~\cite{SteveArko2015}. Thus, pre-calibration is generally an essential component for UAV-RS data processing.

\begin{figure}[tp]
	\centering
	\subfigure[3D physical calibration field]{
		\label{fig:Calibration.1}	
		\includegraphics[height=0.25\linewidth, width=0.43\linewidth]{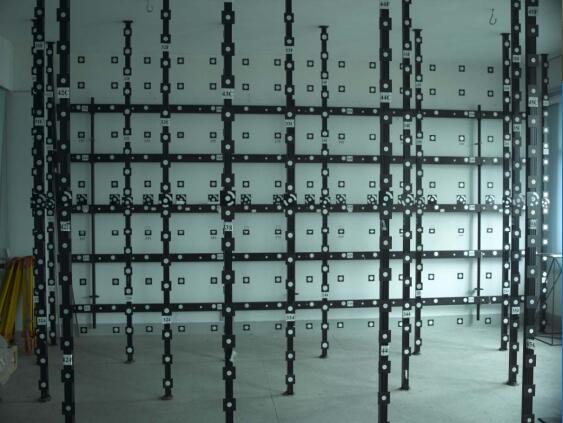}}
	\subfigure[Checkerboard calibration~\cite{Zhang2000}]{
		\label{fig:Calibration.2}
		\includegraphics[height=0.25\linewidth, width=0.43\linewidth]{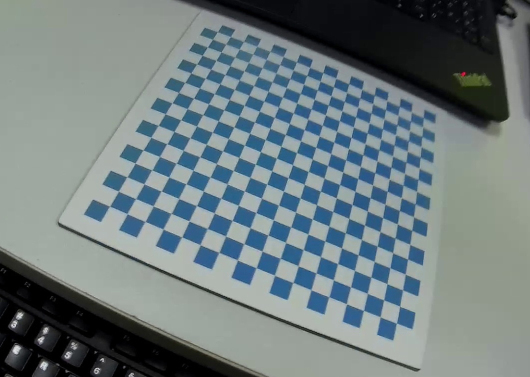}} \\ [-1.5 mm]
	\subfigure[Dual LCD-based method~\cite{Zhan2008}]{
		\label{fig:Calibration.3}
		\includegraphics[height=0.25\linewidth, width=0.43\linewidth]{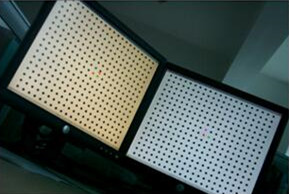}}
	\subfigure[AprilTag-based method~\cite{RichardsonSO2013}]{
		\label{fig:Calibration.4}
		\includegraphics[height=0.25\linewidth, width=0.43\linewidth]{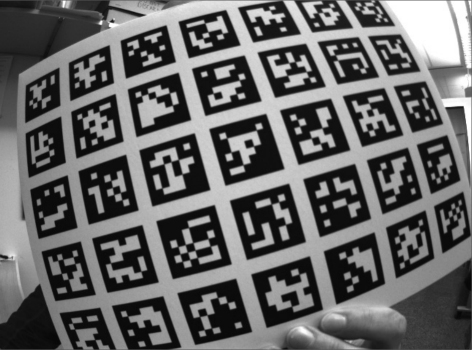}}
	\caption{Examples of camera calibration.}
	\label{fig:RefCalibration}
\end{figure}

In camera calibration, pinhole cameras are often calibrated based on perspective projection model, while fisheye lenses are based on spherical model, orthogonal projection, polynomial transform model, etc.~\cite{KannalaB2006} Methods for camera calibration and distortion correction can be generally classified into two categories: reference object-based calibration and self-calibration. Reference object-based calibration can be performed easily using the projected images of a calibration array, shown in Fig.~\ref{fig:RefCalibration}. The most rigorous method is based on laboratory 3D physical calibration field, where coded markers are distributed in three dimensions with known accurate positions~\cite{TommVirtual2013}. This method provides high-precise calibration parameters, but it is high-cost and inconvenient, and not suitable for frequent recalibration in UAV-RS. An alternative low-cost solution is based on 2D calibration pattern, \emph{e.g.} checker board~\cite{Zhang2000}, completely flat LCD-based method~\cite{Zhan2008} and AprilTag-based method~\cite{RichardsonSO2013}. It has been demonstrated it can achieve the accuracy close to 3D physical calibration field. Different patterns are designed to improve the accuracy and ease of feature detection and recognition under various conditions.

It is worth noting that reference object-based calibration usually requires pre-prepared calibration patterns and extra manual operations, which make it laborious and time-consuming. By contrast, self-calibration, which depends on structural information detected in images without requiring special calibration objects, are more flexible and efficient. It therefore become an active research in recent years, especially for automatic rectification and calibration of fisheye image.

Among these methods,  geometric structures (\emph{e.g.} conics, lines and  plumb lines) are first detected~\cite{WangWZ2008kruppa, HughesDGJ2010, HadiCalUAV2017}. If given at least three conics on distorted image, the camera intrinsic parameters can be obtained from the decomposition of absolute conics. The fisheye image are generally rectified based on the assumption that the straight line should maintain their line property even after the projection of fisheye lens. Several approaches have been proposed to extract geometric structures, such as extended Hough transform~\cite{Bukhari2013} and multi-label energy optimization~\cite{ZhangMleo2015}. However, the effects of rectification are often limited by the accuracy of geometric structure detection. More recently, deep convolutional neural networks (CNNs) based methods have been proposed which tried to learn more representational visual features with CNNs to rectify the distorted image~\cite{RongCalibCNN2016}. In~\cite{YinFisheyenet2018}, an end-to-end deep CNN was proposed which learns semantic information and low-level appearance features simultaneously to estimate the distortion parameters and correct the fisheye image. However, this method does not consider the geometry characteristics, which are strong constrains to rectify distorted images. To this end, Xue~\cite{XueLineRec2019} designed a deep network to exploit distorted lines as explicit geometry constraints to recover the distortion parameters of fisheye camera and rectify distorted image.

\begin{figure}[tp]
	\centering
	\includegraphics[width=0.99\linewidth]{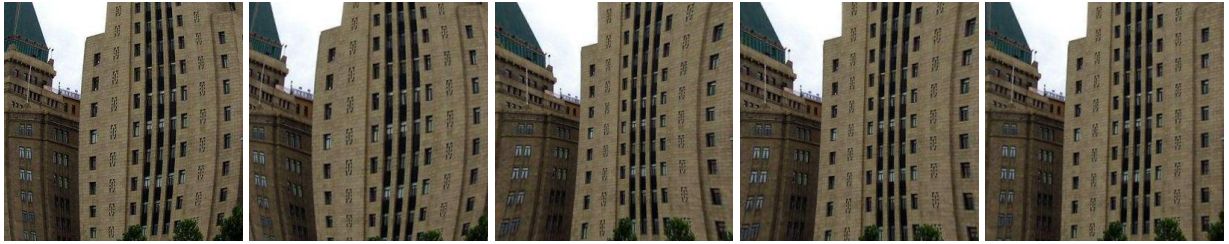}\\
	\includegraphics[width=0.99\linewidth]{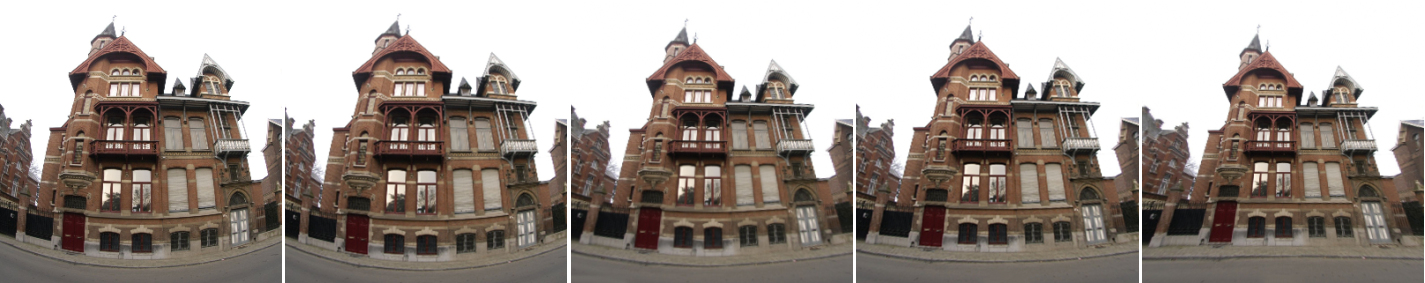}
	\caption{Rectification examples of fisheye image. From left to right are the results by Bukhari~\cite{Bukhari2013}, AlemnFlores~\cite{Alemanipol2014}, Rong~\cite{RongCalibCNN2016} and Xue~\cite{XueLineRec2019}.}
	\label{fig:Calibration}
\end{figure}

Some rectification examples of fisheye image based on self-calibration are shown in Fig.~\ref{fig:Calibration}. The qualitative evaluation on fisheye dataset are reported in Tab.~\ref{tab:fisheyerect}. It can be seen that deep CNNs-based methods (\emph{e.g.} \cite{XueLineRec2019}) achieves the excellent rectification performance for fisheye images.  
However, it remains some challenges need to be solved. The encode of other geometry, such as arcs and plume line, into CNNs is still a challengeable issue. Besides, designing robust geometric feature detection methods especially in case of noises or low texture is also in demand. Another important issue is to improve the accuracy of self-calibration to achieve the comparable accuracy with reference object-based methods. 

\begin{table}[!tb]
	\renewcommand\arraystretch{1.5}
	\caption{Qualitative evaluation of rectification on fisheye image dataset from by Xue~\cite{}, using PSNR, SSIM and reprojection error (RPE).}
	\label{tab:fisheyerect}
	\centering  
	\begin{tabular}{ccccc}
		\hline
		Methods      &Bukhari~\cite{Bukhari2013}  &AlemnFlores~\cite{Alemanipol2014}  &Rong~\cite{RongCalibCNN2016}  &Xue~\cite{XueLineRec2019}\\
		\hline	   
		PSNR          & 11.47             & 13.95            & 12.52         & 27.61 \\
		SSIM           & 0.2429           & 0.3922         & 0.2972       & 0.8746\\
		RPE             & 164.7             & 125.4              & 121.6         & 0.4761\\       
		\hline
	\end{tabular}
\end{table}

\subsection{Combined field of view}

Because of low flight altitude and narrow FOVs of cameras equipped on UAVs, UAV-RS often acquires images with small ground coverage area, resulting in the increase of image amount, flight lines, flight cost and data collection time~\cite{DamianMultiCamera2018}.

\begin{figure}[tp]
	\centering
	\includegraphics[height=0.36\linewidth]{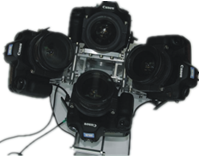}
	\includegraphics[height=0.36\linewidth]{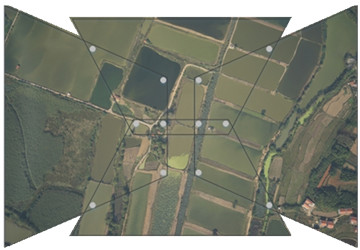} 
	\caption{Left: four-combined camera system in~\cite{LinSu2012}. Right: the overlapping layout of the images projected from the four cameras.}
	\label{fig:Combinationview}
\end{figure}

One alternative solution is the combined wide-angle camera which use multiple synchronized cameras. The images acquired form multi-camera combination system (\emph{i.e.} equivalent large array camera) are rectified, registered and mosaicked to generate a larger virtual image, which can augment the coverage area~\cite{TommVirtual2013}. In contrast to narrow cameras, the combined wide-angle method can increase acquisition efficiency and enlarge the base-height ratio. Besides, it also benefits the image connection especially in some windy conditions. Another advantage is to obtain multi-view images by oblique acquisition, which can overcome dead areas of photograph and sheltered targets. In~\cite{LinSu2012}, the combined wide-angle camera is used for photogrammetric survey and 3D building reconstruction.  Fig.~\ref{fig:Combinationview} shows an example of four-camera system.

The combined wide-angle camera has been well-studied in UAV-RS community. However, it remains challenging to improve acquisition efficiency for larger area mapping. An emerging opportunity is multi-UAV collaboration, which uses fleets of simultaneously deployed “swarming” UAVs to achieve a remote sensing goal. Except for improving spatial coverage and efficiency, multi-UAV collaboration overcome the spatial range limitations of a single platform and thus improve the reliability because of redundancy and allow simultaneous intervention in different places~\cite{WangLi2013, Pajares2015}. Each vehicle can transmit either the collected data or the processed results to ground workstations for further processing or decision. Data can also be shared between different vehicles to guide optimal collaboration. For instance, in~\cite{MerinomultiUAV2012}, a fleet of UAVs, equipped with various sensors (infrared, visual cameras, and fire detectors), cooperated for automatic forest fire detection and localization using a distributed architecture. The heterogeneous sensors increase the complexity of data processing, but they make it possible to exploit the complementarities of vehicles in different locations and flight attitudes and sensors with different perception abilities. Except for multiple UAVs, collaboration can also be performed between UAVs and other remote-sensing platforms, \emph{e.g.} unmanned ground vehicles and unmanned marine surface vehicles~\cite{LinHyyppa2013}.

Multi-UAV collaboration has become an effective means of collecting accurate and massive information and received increased attention recently. It has been widely used in commercial performance, but it is noting that there are some reports about accidents of multi-UAV systems. There is still a long way to go for broad applications of multi-UAV systems in remote-sensing community. Some problems are worth the effort, such as system resilience, complexity and communication between the UAVs, navigation and cooperative control in harsh conditions, environmental sense and collision avoidance, detection of anomalies within the fleet and disruption handling including environmental obstacles, signal interference and attack~\cite{TommVirtual2013,Barmpounakis2016UAVTransp}. Besides, how to configure the number of UAVs and plan flight routes to achieve optimal efficiency and performance is also a challenging issue~\cite{Riehl2009Cooperative, Lim2018MultiUAV}.

\subsection{Low-altitude UAV image matching}
\label{Sec:Match}

Image matching is one of the fundamental technologies in photogrammetry and computer vision, which is widely used in image registration, image stitching, 3D reconstruction, etc.~\cite{Zhang2017,ZhuoKoch2017,JiangJiang2018}. It is a long-standing and challenging task, especially for UAV images, due to the strong geometric deformations (\emph{e.g.} affine distortion), viewpoint changes, radiation/illumination variances,  repetitive or low texture and occlusion. Although numerous matching algorithms have been proposed~\cite{Gruen2012} over the last decades, they may be fail to provide good performance for low-altitude UAV images.

\begin{figure}[tp]
	\centering
	\subfigure[Matching nadir and oblique images~\cite{XiaoGuo2016}]{
		\includegraphics[width=0.99\linewidth]{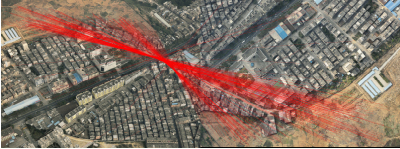}}\\
	\subfigure[Matching ground to aerial images~\cite{Zhou2017Progressive}]{
	    \includegraphics[width=0.99\linewidth]{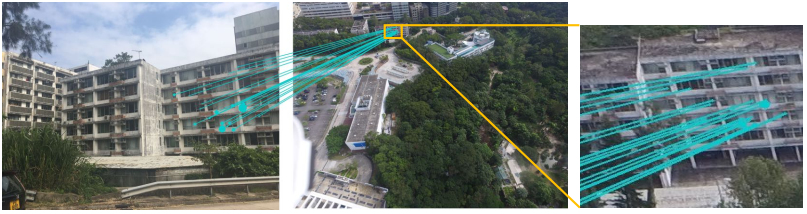}}\\
	\subfigure[Matching UAV image to geo-reference images~\cite{ZhuoKoch2017}]{
		 \includegraphics[width=0.99\linewidth]{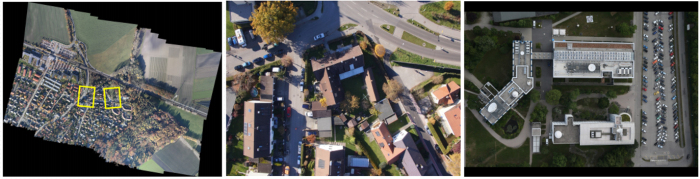}}
	\caption{Low-altitude UAV image matching.}
	\label{fig:Matching}
\end{figure}

\vspace{2mm}
\subsubsection{\textbf{Multi-view image matching}}
Multi-view photography can acquire data from nadir and side-looking directions, especially in UAV-based oblique photogrammetry. However, this special data collection manner makes image matching astonishingly difficult, \emph{e.g.} vertical and oblique image matching, because of the obvious difference in their appearances caused by the wide baseline and large viewpoint changes, especially affine deformations~\cite{HuZhu2015}. 

Some attempts have been made to create local descriptors invariant to affine distortions, such as maximally stable extremal region (MSER), Harris/Hessian-affine and affine-SIFT (ASIFT), MODS~\cite{Wang2018unmanned}. Although they can handle images with viewpoint variances, they either provide small amount of correspondences or suffer from excessive time consumption and memory occupancy. Besides, these methods are not designed specifically for UAV cases, and may have difficulty in meeting the demands of even distribution of correspondences in images with uneven distributed texture.

There are usually two strategies proposed to handle affine deformations in UAV image matching. One is to perform multi-view image matching based on MSER. The local regions are often normalized to circular areas, on which interest points are selected and matched. Considering the small quantity and uneven distribution of matching pairs, some geometric constraints, \emph{e.g.} local homography constraint, can be used to guide the propagative matching~\cite{YuYDY2018}. 
The other is to apply geometric rectification before image matching~\cite{JiangJiang2017}. If images collected by UAVs contain rough or precise exterior orientation elements and camera installation parameters, they can be used for geometric rectification of oblique UAV images to relieve perspective deformations. With the conventional descriptor matching methods, sufficient and well-distributed tie-points are then extracted and matched. The oblique images can also be rectified by coarse initial affine-invariant matching~\cite{Wang2018unmanned}. To achieve reliable feature correspondence, spatial relationships and geometrical information can be adopted to guide matching process and remove outliers, \emph{e.g.} local position constraint, cyclic angular ordering constraint and neighborhood conserving constraint in~\cite{HuZhu2015}.

To obtain the matching pairs as evenly distributed as possible, the divide-and-conquer and the tiling strategy are often adopted~\cite{JiangJiang2017}. Images are split into blocks, and features are extracted and matched from the corresponding blocks. The number of points in each block can be adaptively determined by information entropy~\cite{SUNL2SIFT2014, AiHu2015}.

Although significant progresses have been achieved in UAV multi-view image matching, there is still plenty of room for improvement. Due to the powerful ability of feature representation of deep CNNs and huge success in image classification and target detection~\cite{ZhangZD2016}, deep learning shows explosive increase in image matching recently~\cite{Balntas2017}. Deep neural networks are designed to learn a local feature detector, such as temporally invariant learned detectors from pre-aligned images of different time and seasons~\cite{VerdieYFL2015}, covariant local feature detector which regards the feature detection as a transformation regression problem~\cite{zhangYu2017}. In fact, limited progresses have been made in deep feature detection, due to the lack of large-scale annotated data and the difficulty to get a clear definition about keypoints. By contrast, great efforts have been made on developing learned descriptors based on CNNs, which have obtained surprising results on some public dataset. Feature descriptors are often developed by Siamese or triplet networks with well-designed loss functions, such as hinge loss, SoftPN, joint loss and global orthogonal regularization~\cite{ZhangYKC2017}. Besides, some geometric information are integrated to facilitate local descriptor learning, \emph{e.g.} patch similarity and image similarity in~\cite{LuoGeoDesc2018}. In~\cite{Altwaijry2016}, image matching is considered as a classification problem. An attention mechanism is exploited to generate a set of probable matches, from which true matches are separated by a Siamese hybrid CNN model.

However, it is well-known that deep learning-based image matching requires large annotated datasets, while the existing datasets are often small or lack of diversity. The limited data source reduces the generalization ability of deep models, which may causes poor performance compared with hand-crafted descriptors~\cite{LuoGeoDesc2018}. Although a diverse and large-scale dataset HPatches has been released recently, it is not constructed from UAV-RS images.

\vspace{2mm}
\subsubsection{\textbf{Matching with non-UAV images}}
UAV images are often co-registered with existing georeferenced aerial/satellite images to locate ground control points for spatial information generation, UAV geo-localization~\cite{Castillo2017AutoMetric}. To increase the number of keypoints, the boundaries of super-pixels are adopted as feature points, followed by one-to-many matching scheme for more matching hypotheses~\cite{ZhuoKoch2017}. Geometric constraints based on pixel distance to correct matches are employed for mismatch removal at repetitive image regions. Considering variation of illumination between UAV and satellite images, illumination-invariant image matching is proposed based on phase correlation to match the on-board UAV image sequences to a pre-installed reference satellite images for UAV localization and navigation~\cite{WanLiu2016}.

It is a huge challenge that matching UAV images with ground/street-view images due to the drastic change in viewpoint and scales that causes the failure of traditional descriptor-based matching. Some approaches attempted to warp the ground image to the aerial view to improve feature matching~\cite{QiWu2014}. Besides, in~\cite{WolffCollins2016}, the matching problem is considered as a joint regularity optimization problem, where the lattice tile/motif is used as a regularity-based descriptor for facades. Three energy terms, \emph{i.e.} edge shape context, Lab color features and Gabor filter responses, are designed to construct matching cost function. Another promising method is to employ the CNN to learn representations for matching between ground and aerial images. In~\cite{Hu2018CVMNet}, a cross-view matching network was developed to learn local features and then form global descriptors that are invariant to large viewpoint change for ground-to-aerial geo-localization. In addition, to handle image matching across large scale differences, which include small-scale features to establish correspondences, Zhou,~\emph{et al.}~\cite{Zhou2017Progressive}  divided the image scale space into multiple scale levels and encoded it into a compact multi-scale representation by bag-of-features. The matching then restricts the correspondence search of query features within limited related scale space, and thus improve the accuracy and robustness of feature matching under large scale variations.

\vspace{2mm}
\subsubsection{\textbf{Challenges in UAV image matching}}
Though tremendous efforts have been devoted to low-altitude image matching, there are many problems need to consider, as follows. 
\begin{itemize}
	\item Except for interest points, geometric structure features which represent more information, \emph{e.g.} lines, junctions, circles and ellipse,  can also play a significant role in multi-view image matching, especially in urban scenarios~\cite{Xia2014, XueXia2018, Xue2019}. Geometric features often have invariant to radiometric change and scene variation over time. A small amount of work is concentrated on line-based image matching~\cite{ZhaoGoshtasby2016}. More efforts are worth to develop image matching based on geometric features.
	\item Deep learning-based image matching is a promising method for UAV image matching. However, the lack of large-scale annotation datasets from UAV data hinders development of novel and more powerful deep models. Geometric information (\emph{e.g.} local coplanar) are often overlooked in learning process, which can be encoded into deep neural networks to improve matching performance. Besides, except for feature detection and description,  geometric verification can also be encoded into neural networks for outlier rejection~\cite{Yi2017LearningTF}. Moreover, how to learn detector and descriptor of structure feature by CNNs for image matching is also a challenge.
	\item Cross-view image matching has drawn a lot of attention in recent years. They play important roles in image-based geo-localization and street-to-aerial urban reconstruction. However, large viewpoint/scale differences should be well considered. More powerful deep models or more effective scale-space image encoding approaches are in demand.
\end{itemize}

\begin{table*}
	\renewcommand\arraystretch{1.2}
	\centering
	\caption{Comparison of three SfM paradigms$^1$.} 
	\label{Tab:SfM}
	\begin{threeparttable}	
		\begin{tabular}{p{0.35\columnwidth}<{\centering}|p{0.56\columnwidth}<{\centering}|p{0.7\columnwidth}<{\centering}|p{0.25\columnwidth}<{\centering}}
			\hline
			Item                                                  & Incremental                       & Global                     & Hierarchical \\ \hline		
			Match graph initialization                & Initialized by selected seed image pairs        
			& All images are treated equally
			& Atomic models		\\\hline
			Camera registration                          & Perspective-n-Point (PnP), 2D-3D correspondences 
			& Rotation and translation averaging
			& 3D-3D fusion	\\	\hline
			Bundle adjustment                            & Iterative, many times 
			& One time
			& BA when merging\\\hline
			Advantages                                       & Robust, high accuracy, good completeness of reconstructed scene
			& Evenly-distributed errors, high efficiency
			& Fewer BA steps\\\hline
			Disadvantages                                   & Prone to drifting errors, low efficiency
			& Prone to noisy pairwise	matches, relatively low accuracy, low completeness of reconstructed scene
			& Model merging, graph partition	\\\hline
			Tools                                                & Bundler, OpenMVG, VSFM, MVE, ColMap 
			& OpenMVG, 1DSfM, DISCO, Theia  
			& Research papers		\\ \hline
		\end{tabular}
		\begin{tablenotes}
			\footnotesize
			\item[$^1$] Refer to: Tianwei Shen, Jinglu Wang, Tian Fang, Long Quan, Large-scale 3D Reconstruction from Images, ACCV tutorial, 2016.
		\end{tablenotes}
	\end{threeparttable}
\end{table*}

\subsection{Low-altitude automatic aerial triangulation}	

Aerial triangulation, namely recovering camera poses and 3D structures of scene from 2D images, is a fundamental task in photogrammetry and computer vision. For manned aerial photogrammetry that collects images vertically, automatic aerial triangulation (AAT) has been well-studied~\cite{Forstner2016}. As to UAV-based photogrammetry, however, it is demonstrated that the long established and proven photogrammetric AAT cannot handle UAV blocks~\cite{QinGrun2013}. This is because low-altitude UAV-RS breaks the acquisition mode of traditional photogrammetry (discussed in~\ref{Sec:Challenges}) and does not meet the assumptions of conventional AAT~\cite{ZhangXiong2011}.

In the last few years, structure from motion (SfM) brings the light to low-altitude UAV AAT~\cite{Westoby2012}. SfM estimates the 3D geometry of a scene (structure), the poses of cameras (motion) and possibly camera intrinsic calibration parameters simultaneously, without need either camera poses or GCPs to be known prior to scene reconstruction~\cite{ozyesil2017}.
Some tests that apply SfM software for UAV-based aerial triangulation have demonstrated that SfM can break through the obstacles of UAV irregular blocks for robust low-altitude UAV AAT~\cite{Colomina2014}.

\vspace{2mm}
\subsubsection{\textbf{Structure from motion}}
SfM is generally divided into three types: incremental, global and hierarchical SfM, according to their initialization ways of camera pose. A simple comparison of these three SfM paradigms can be seen in Tab.~\ref{Tab:SfM}. Besides, to make full use of incremental and global SfM, hybrid SfM is proposed to estimate camera rotations in a global way based on an adaptive community-based rotation averaging, and estimate camera centers in an incremental manner~\cite{CuiGao2017}. To achieve city-scale sparse reconstruction, Zhu, \emph{et al.}~\cite{ZhuSZZWFQ2017} grouped cameras and performed local incremental SfM in each cluster, and then conducted global averaging between clusters. The hybrid SfM method possesses both robustness inheriting from incremental manner and efficiency  inheriting from global manner. However, repeated BA is still needed in estimation of camera centers, which needs more efforts.

Recently, the semantic information is integrated into sparse reconstruction~\cite{BaoSemanticSFM2012}. These methods consider the semantic SfM as a max-likelihood problem to jointly estimate semantic information (\emph{e.g.} object classes) and recover the geometry of the scene (camera pose, objects and points). However, due to their large memory and computational cost, this manner is often limited to small scenes and low resolution. Besides, semantic information can also be used to constrain feature matching and bundle adjustment by semantic consistency~\cite{ChenSFMData2018}.

\vspace{2mm}
\subsubsection{\textbf{Image orientation}}
In SfM, camera poses are often estimated from feature correspondences by solving the perspective-n-point problem and then optimized by BA. Besides, external orientation sensors can be adopted for camera pose estimation. If UAVs equip with high-quality GPS/IMU, positions and orientations of cameras can be estimated from GPS/IMU data directly without the need of GCPs, namely direct sensor orientation or direct georeferencing~\cite{TurnerLucieer2014}. Besides, orientation parameters from GPS/IMU can be used to initialize the camera poses, and then integrate them into aerial triangulation for bundle adjustment, \emph{i.e.}, integrated sensor orientation. However, UAVs are often mount with low-accuracy navigation sensors, due to payload limitation and high costs of low-weight and high-precise navigation systems. Therefore, ground control points are adopted for high-precise aerial triangulation, called indirect sensor orientation, which is time-consuming and laborious.

The existing SfM approaches generally heavily rely on accurate feature matching. Some failure may be caused by low/no texture, stereo ambiguities and occlusions, which are common in natural scenes. Thus, to break through these limitations, deep models are applied for camera pose estimation or localization recently~\cite{yin2018geonet}. In~\cite{Kendall2015}, a PoseNet is designed to regress the camera pose from a single images in an end-to-end manner. Besides, the traditional SfM is modeled by learning the  monocular depth and ego-motion in a coupled way, which can handle dynamic objects by learning a explain-ability mask~\cite{zhou2017unsupervised, Vijay2017SfMNet}. However, the accuracy of these methods is far from that of traditional SfM. Besides, they are dependent on data set and are difficult to provide good generalization capabilities. To build more diverse data sets and encode more geometric constraints into deep models are worth efforts.

\begin{figure*}[!tb]
	\centering
	\includegraphics[width=0.98\linewidth]{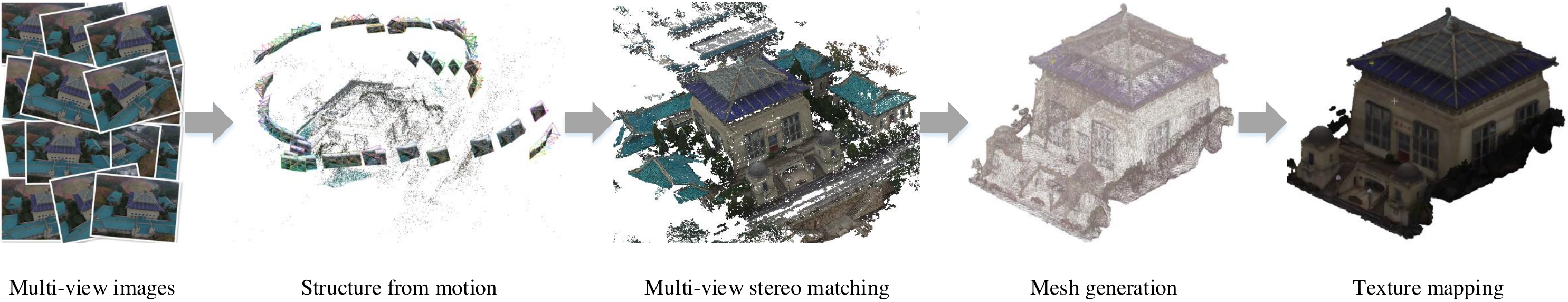}
	\caption{Image-based multi-view 3D reconstruction. Based on UAV images, SfM is performed to estimate camera poses and sparse 3D structure. Dense reconstruction is then adopted to generate dense 3D scene structure. Surface reconstruction is conducted to generate a surface model. After texture mapping, the real 3D model is reconstructed.}
	\label{fig:3DRecons}
\end{figure*}

\vspace{2mm}
\subsubsection{\textbf{SfM for Rolling Shutter Cameras (RSC)}}
Most off-the-shelf cameras are equipped with a rolling shutter due to the low manufacturing cost. However, its row-wise exposure delay bring about some problems. In the presence of camera motion, each row is captured in turn and thus with a different pose, which causes severe geometric artifacts (\emph{e.g.} skew and curvature distortions) in the recorded image. This defeat the classical global shutter geometric models and result in severe errors in 3D reconstruction. Thus, new methods adapted to RSC are strongly desired.

Some works contribute to correct rolling shutter distortions~\cite{RengarajanRSC2016}. One way is to use inter-frame correspondences to estimate the camera trajectory and register frames. The continuity and smoothness of camera motion between video frames can also be combined to improve performance. Another way is to perform correction as an optimization problem based on straightness, angle, and length constraints on the detected curves to estimate the camera motion and thus rectify the rolling shutter effect. This method is sensitive to feature choice and extraction. Recently, CNNs are adopted to automatically learn the interplay between scene features and the row-wise camera motion and correct the distortions~\cite{RengarajanRSC2017}. Large scale data sets are obviously required. They often train CNNs on synthetic dataset which may be different from the real cases, but it is a promising method.

Rolling shutter effects are modeled in the case of conventional SfM~\cite{KlingnerRSC2013, ImRSC2015}. The complex RSC model is shattered into a constellation of simple, global-shutter, linear perspective feature cameras. The poses (\emph{i.e.} rotation and translation) of each feature are linearly interpolated according to their vertical position in the image between successive key poses. Usually, a linear interpolation is used for translation and a spherical linear interpolation is used for rotation. In general, one may insert as many key poses as the tracked features.

\vspace{2mm}
\subsubsection{\textbf{Challenges in aerial triangulation}}
Although aerial triangulation/SfM is a long-standing problem, it still faces many challenges, such as very large-scale and high-efficiency SfM, AAT with arbitrary images, multi-source data (ground/street images and UAV images) AAT. Besides, there is a long way to go for semantic SfM and deep CNNs for camera pose estimation.

\subsection{Dense reconstruction}

A complete workflow of 3D construction includes structure-from-motion, dense reconstruction, surface reconstruction and texture mapping~\cite{NexRemondino2014}, shown in Fig.~\ref{fig:3DRecons}. Once a set of UAV images are oriented, namely known camera poses, the scene can be densely reconstructed by dense image matching (\emph{i.e.} multi-view stereo matching), the focus of this section.

\vspace{2mm}
\subsubsection{\textbf{Multi-view stereo (MVS) Reconstruction}}
Numerous multi-view stereo algorithms have been proposed, \emph{e.g.} semi-global matching, patch-based methods, and visibility-consistent dense matching~\cite{RemondinoS2014}. To search for correspondences, similarity or photo-consistency measures are often adopted to compare and estimate the likelihood of two pixels (or groups of pixels) in correspondence. The most common photo-consistency measures include normalized cross correlation, sum of absolute or squared differences, mutual information, census, rank, dense feature descriptors, gradient-based algorithms and bidirectional reflectance distribution functions~\cite{FurukawaC2015}. MVS is often formulated as a function of illumination, geometry, viewpoints and materials, and thus can be regarded as a constrained optimization problem solved by convex optimization, Markov random fields, dynamic programming, graph-cut or max-flow methods~\cite{RemondinoS2014}.

Most conventional multi-view stereo matching methods are adopted directly for UAV image-based surface reconstruction~\cite{HarwinLucieer2012}. 
Considering the perspective distortions in oblique images, epipolar rectification is performed based on cost of angle deformation before MVS matching~\cite{LiuGuo2016}. To minimize the influence of boundary, a hierarchical and adaptive phase correlation is adopted to estimate disparity of UAV stereo images~\cite{LiLiu2016}. Besides, some tricks are proposed to improve the performance of conventional methods, including graph network, image-grouping and self-adaptive patch~\cite{XiaoGuo2016}.

\vspace{2mm}
\subsubsection{\textbf{Learning-based MVS}}
The aforementioned methods use hand-crafted similarity metrics and engineered regularizations to compute dense matching, and are easily affected by sudden changes in brightness and parallax, repeated/no textures, occlusion, large deformations, etc. 

Recent success on deep learning research has attracted interest to improve dense reconstruction. Numerous works apply CNNs to learn pair-wise matching cost~\cite{ZbontarMCNet2016} and cost regularization~\cite{SekiSGMNet2017}, and also perform end-to-end disparity learning~\cite{KendallGCNet2017}. However, most methods focus on stereo matching tasks, which are non-trivial to extend them to multi-view scenarios. Furthermore, the extended operations do not fully utilize the multi-view information and lead to less accurate result. Besides, input images could be of arbitrary camera geometries.
 
There are fewer works on learned MVS approaches. SurfaceNet~\cite{JiSurfaceNet2017} and Learned Stereo Machines~\cite{KarLSM2017} encode camera information in the network to form the cost volume, and use 3D CNN to infer the surface voxels. However, these methods are limited by huge memory consumption of 3D volumes and thus only handle small-scale reconstructions. Thus, DeepMVS~\cite{HuangDeepMVS2018} takes a set of plane-sweep volumes for each neighbor image as input and  produces high-quality disparity maps, which can handle an arbitrary number of posed images. MVSNet~\cite{YaoMVSNet2018} builds 3D cost volume upon the camera frustum instead of the regular Euclidean space and produces one depth map at each time. Thus, this approach  makes large-scale reconstruction possible. However, due to the annotated data without the complete ground truth mesh surfaces, this method may be deteriorated by occluded pixels. The works in~\cite{LiuEvaluation2018} provides comparison experiments and demonstrates that deep learning based methods and conventional methods perform almost the same level. While deep learning based methods have better potential to achieve good accuracy and reconstruction completeness.

\vspace{2mm}
\subsubsection{\textbf{Challenges in dense reconstruction}}
Although great success has been achieved, there remains some challenges which need more efforts, as follows.

\begin{itemize}
	\item \emph{Specular object reconstruction.} Most MVS algorithms often impose strong Lambertian assumption for objects or scenes, however, there are many specular objects or isotropic reflectance objects in man-made environments. Multi-view reconstruction of these glossy surfaces is a challenging problem. One promising method may be to adopt generative adversarial network for transferring multiple views of objects with specular reflection into diffuse ones~\cite{WuMVS2018}.
	
	\item \emph{Dynamic scene modeling.} Most existing 3D reconstruction methods are under the assumption of a static rigid scene. How to reconstruct dynamic scene is a challenging issue. One possible way is to pre-segment the scene into different regions where is locally rigid and apply rigid SfM and MVS to each of the regions~\cite{KumarDL2017}. 
	
	\item \emph{Multi-source 3D data fusion.} Few attempts have been carried out in the fusion of aerial and ground-based 3D point clouds or models~\cite{GaoHCSH2018}. The large differences in camera viewpoints and scales impose a tricky issue to the alignment of aerial and ground 3D data. Moreover, it is also a difficult task to reconstruct a single consistent 3D model that is as large as an entire city with the details as small as individual objects. 

\end{itemize}

\subsection{Image stitching}

\begin{figure}[tbp]
	\centering
	\subfigure[Ortho-rectification based stitching. Left: inaccurate mosaic map generated by the direct georeferencing using the original inaccurate IMU/GPS data. Right: mosaic map generated based on registration with the reference map in ~\cite{FarajiQi2016}.]{
		\includegraphics[height=0.45\linewidth]{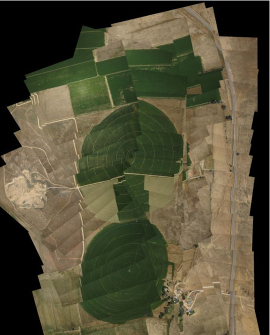} 
	    \includegraphics[height=0.45\linewidth]{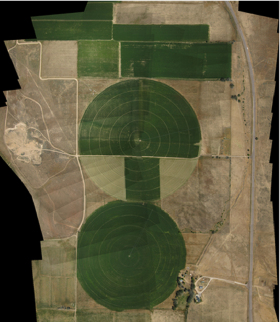} }
	\subfigure[Transformation based stitching. Automatically constructed urban panorama with 14 wide-baseline images based on mesh-optimization stitching method proposed in~\cite{ZhangHC16Multi}. ]{
		\includegraphics[width=0.8\linewidth]{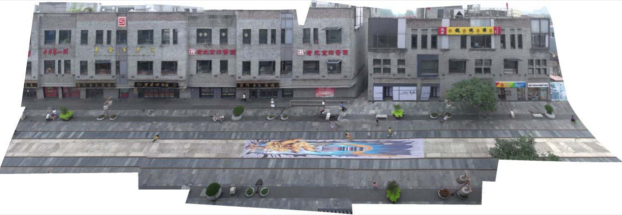} }
	\caption{Examples of image stitching.}
	\label{fig:Stitch}
\end{figure}

Due to the small footprint of UAV images, it is essential to develop automatic image stitching/mosaicking techniques to combine multiple images with overlapping regions into a single large seamless composite image with wide FOV or panorama~\cite{XiangImage2016, Xianglocally2016}. 
Image stitching generally includes geometric correction and image composition. Images acquired from different positions and attitudes are registered to an identical mosaic plane or reference plane in geometric correction, and then the inconsistencies in geometry and radiation (\emph{e.g.} color or brightness) among geometric-corrected images are mitigated or eliminated by image composition. Some examples of image stitching are shown in Fig.~\ref{fig:Stitch}. According to the different methods for geometric correction, image stitching can be divided into ortho-rectification based stitching and transformation based stitching, detailed below. Image composition, including seamline generation, color correction and image blending, is generally similar to that of other remote-sensing platforms. Recognizing space limitations, we therefore refer interested readers to several papers~\cite{LiHui2015,TianLi2016,SongJia2017} for the detailed description.

\vspace{2mm}
\subsubsection{\textbf{Ortho-rectification based image stitching}}
Ortho-rectification based image stitching is the essential step for generation of digital orthophoto maps, which are used for photogrammetric recording and document and are also the base map for remote sensing interpretation. Images are often ortho-corrected based on camera poses and 3D terrain information, (\emph{e.g.} DEM/DSM and GCPs), to reduce the geometric deformation and achieve spatial alignment on the same geographical coordinate system. In~\cite{TurnerLucieer2014}, DEMs/DSMs are generated from SfM point clouds, which are then transformed into real-world coordinates based on direct/indirect/integrated georeferencing. In~\cite{AndreaMark2011}, images are corrected by global transformations derived from the relationships between GCPs and the corresponding image points. Considering the inaccuracy of exterior orientation from GPS/IMU and the difficulties in acquisition of GCPs, another way for ortho-rectification is based on registration with the aerial/satellite orthorectified map~\cite{FarajiQi2016}. In contrast, this way is more efficient and convenient due to the avoidance of complex aerial triangulation and DEM generation as well as the laborious acquisition of GCPs, however, its mandatory prerequisite is the reference maps.

\vspace{2mm}
\subsubsection{\textbf{Transformation based image stitching}}
Ortho-rectification based image stitching can rectify the geometric distortions and provide geographic coordinate information, however, it is generally computation-complex and time-consuming, which make it unsuitable for time-critical remote-sensing applications~\cite{BangKim2017}, such as disaster emergency and security monitoring. This approach provides an effective mosaic method based on transformations calculated from matching correspondences between adjacent images~\cite{YuWYYX2018}.

A simple approach is to exploit one global transformation to align images~\cite{XiaYao2017}. However, it only works well under the assumptions of roughly planar scenes or parallax free camera motion~\cite{Zhang2017}, which may be violated in most UAV-based data acquisition cases. Though advanced image composition can mitigate stitching artifacts generated by these methods, they remain when there are misalignments or parallax. 

To this end, spatially-varying warping methods have been proposed for image alignment. One is to adopt multiple local transformations to locally align images, including as-projective-as-possible warping~\cite{ZaragozaChin2014} and elastic local alignment model~\cite{LiELA2018}. The other is to consider registration as an energy optimization problem with geometric or radiometric constraints based on mesh optimization model~\cite{XuOu2016, ZhangHC16Multi}. Local transformations can also be integrated with mesh models to provide good stitching~\cite{XiangXBZ17}. Spatially-varying warping models can handle moderate parallax and provide satisfactory stitching performance, but it often introduces projective distortions, \emph{e.g.} perspective and structural distortions, due to the nonlinear of these transformations. Some methods have been proposed to handle distortions, such as global similarity prior model~\cite{ChenYS2016gsp}, structural constraint model~\cite{XiangXBZ17}, but more efforts should be put into stitching images accurately with reduced distortion.

Another approach is seam-guided image stitching~\cite{LinJiang2016}, which hold potential for handling large parallax. Multiple transformation hypotheses can be estimated from different groups of feature correspondences. Seam-line quality is then adopted to evaluate the alignment performance of different hypotheses and select the optimal transformation. This approach adopts a local transformation for global alignment, thus it would get trapped when handling images with complex multi-plane scenes.

\vspace{2mm}
\subsubsection{\textbf{Challenges in image stitching}}
Although numerous stitching methods have been developed, it is also an  open problem, especially for stitching images with efficiency, registration accuracy and reduced distortion. More works should be devoted into high-efficiency/real-time image stitching, large-parallax image stitching and distortion handling in the future. Besides, there exists some attempts of deep learning in homography estimation and image dodging recently~\cite{GuoPan2017,NguyenHomoNet2018}. However, there is still a lot of room for improvement. It is a promising and worthwhile direction in image stitching.

\subsection{Multi-sensor data registration}

With the advent of increasing available sensors, UAV-based remote sensing often equip with multiple remote-sensing sensors (\emph{e.g.} visible cameras, infrared sensors or laser scanners), which can either collect a variety of data at a time to achieve multiple tasks or integrate these complementary and redundant data for better understanding of the entire scene. However, the data from multiple sensors often have dramatically different characteristics, \emph{e.g.} resolution, intensity, geometry and even data dimension, due to different imaging principles. This imposes a huge challenge to how to integrate multi-sensor data for remote sensing applications~\cite{NagaiChen2009}.

Multi-sensor data registration is a mandatory prerequisite.  Multi-sensor data is then fused for data interpretation. Due to limitations on space, this section focus on multi-sensor data registration. Remote-sensing data fusion will not be discussed here and can be referred to the surveys~\cite{SchmittZhu2016,XiangFusion2015}. 

\vspace{2mm}
\subsubsection{\textbf{Multi-band image registration}}
The registration of multi-band images, \emph{e.g.} visible and infrared images, visible and SAR images, has caused great concern in recent years. The area-based method commonly adopts intensity statistical information to handle the large appearance differences, such as mutual information and entropy-based measures~\cite{KimLee2008}. Considering its difficulties to handle large radiometric distortions because they are mainly based on image intensities, structure features, which are more robust to radiometric changes, are integrated as similarity metrics to improve registration performance, such as gradient, edge information, local self-similarity and phase congruency~\cite{YeSBS2017}. However, these methods are computationally expensive. 

\begin{figure}[]
	\centering
	\subfigure[Recognition rate of different matching methods]{
		\includegraphics[width=0.9\linewidth]{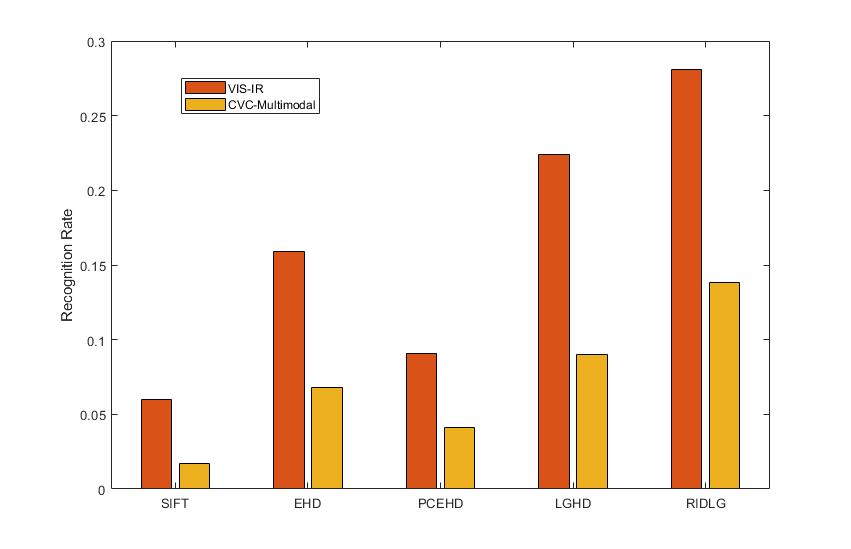}} \\
	\subfigure[Recognition rate of different rotations]{
		\includegraphics[width=0.9\linewidth]{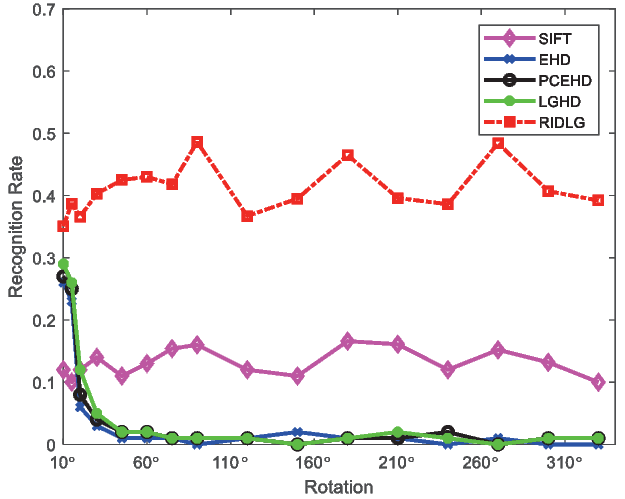}}
	\caption{Visible and infrared image matching in~\cite{ChenDEO2018}. (a) the average recognition rate of different multi-modal image matching methods. (b) the recognition rate of different rotations. This experiments are conducted on VIS-IR and CVC-Multimodal datasets. Recognition rate is defined as the number of correct matches among all the correspondences. SIFT: scale-invariant feature transformation; EHD: edge histogram descriptor; PCEHD: phase congruency and edge histogram descriptor; LGHD: Log-Gabor histogram descriptor; RIDLG: rotation invariant feature descriptor based on multi-orientation and multi-scale Log-Gabor filters. Figure (a) demonstrates the effectiveness of methods based on structure information. However, most methods provide poor performance under rotation issues (Figure (b)). Thus, there is still plenty of room for improvement.}
	\label{fig:compareregis}
\end{figure}

The feature-based registration often extracts geometric features and then matches them based on descriptor matching~\cite{HanPauwels2013,YahyanejadR2014}. However, traditional gradient/intensity based feature descriptors are not suitable for multi-modal image matching due to the large gradient differences. Thus, some structure features, \emph{e.g.} line segments and edges, are described by geometrical relationship, edge histogram or Log-Gabor filters~\cite{ChenDEO2018}. Fig.~\ref{fig:compareregis} shows some promising results and demonstrates the effectiveness of description based on structure information, but they are far from satisfactory performance. Therefore,  there still exists great space for further development. Besides, it is challenging to extract highly repeatable homonymy features from multi-band images because of non-linear radiometric differences.

\vspace{2mm}
\subsubsection{\textbf{Registration of LiDAR and optical images}}
It is a common case in UAV-based remote sensing. The simple way is direct georeferencing, however, it is difficult to achieve high-accuracy registration due to vibration of platforms, unknown exposure delay, limitations of hardware synchronization and calibration, low accuracy of onboard GPS/IMU sensors. There are often three other strategies as follows.
  
\begin{itemize}
	\item The problem can be considered as a multi-modal image registration, by transforming LiDAR data into images, including  grayscale-encoded height and return-pulse intensity images (also called reflectance images). Thus, area-based and feature-based multi-modal image registration can be adopted. 
	\item The problem can be converted as the registration of two point-sets: LiDAR point set and image-derived point set. The iterative closest point (ICP) algorithms can be used. Salient features are often extracted from two point-sets for registration, used as the initialization of ICP~\cite{YangChen2015}. 
	\item Registration can also be performed between LiDAR point cloud and optical images directly, often based on line and plane features. 
\end{itemize}

In the first method, area-based methods are often affected by return-pulse intensity calibration, which determines the quality and correctness of intensity image.  In contrast, feature-based methods provide robust registration~\cite{LiuTong2016}. The transformation error may be another issue that affects registration. In the second method, there is a big difference between two point-sets. LiDAR provides a set of irregularly distributed points with abundant information along homogeneous area but poor information along object space discontinuities, while the image-derived point set is the opposite. Besides, the accuracy of image-derived point set and the initialization of ICP are also non-trivial issues. In the third method, it may be a challenging task to find conjugate features automatically in both datasets.

\vspace{2mm}
\subsubsection{\textbf{Challenges in data registration}} 
Multi-sensor data registration has gained increasing attentions, and there are challenges need to be devoted. Considering the invariance of semantic information of targets in multi-modal images, the semantic feature or target can be extracted for registration. Besides, few works are devoted to consider complex cases with scale, rotation and affine issues in multi-modal image registration. Moreover, multi-sensor image registration based on CNNs is a promising direction.

\subsection{High-performance data processing}

With large amount of data,  the complexity of processing algorithms and the request for a fast response, the time to process and deliver the remote-sensing products to users becomes a main concern for UAV-RS. Consequently, automatic and efficient processing has become a key challenge for UAV data processing.

One available way is to perform data processing with low-complexity algorithms and few manual intervention, such as image location estimation with less/no GCPs or direct georeferencing~\cite{TurnerLucieer2014}. In deep CNNs, some tricks for light-weight models are proposed, including removing regions of proposal for object detection~\cite{TijtgatRanst2017}, model compression and acceleration by parameter sharing, pruning, low-rank matrix decomposition and knowledge distillation~\cite{Cheng2017}.

Another effective solution is high performance computing (HPC)~\cite{LeeHPC2011,GhamisiHSI2017}, such as parallel computing. Unlike serial computation for data processing, parallel computing allows the simultaneous use of multiple computer resources to accelerate data processing. Some available strategies are as follows.
 
 \begin{itemize}
	\item \emph{Hardware accelerators}, including field-programmable gate array (FPGA) and graphical processing unit (GPU). GPU holds great potential in computer intensive, massive-data-parallel computation and has gained lots of attentions for UAV data processing~\cite{HongTong2015, ChenHPC2015}. They can also be used for on-board real-time processing.
	\item \emph{Cluster computers.} The processing task should be broken down into subtasks and then allocated to different computers. It is particularly appropriate for efficient information extraction from very large local data archives.
	\item \emph{Cloud computing.} It is a sophisticated high-performance architecture and used for service-oriented and high-performance computing. For instance, cloud computing are used for processing image data to generate 3D models in distributed architectures~\cite{ZhangBA2018}.
\end{itemize}

For large-scale data acquisition of UAV-RS, it may be challenging that how to achieve optimal path planning to collect the optimal and minimum data to meet the requirements of remote sensing tasks, so as to reduce invalid or redundant data, and mitigate the difficulty of extracting information from massive data. Another important challenge related to fast computing is the volume, weight, cost and high energy consumption of high-performance computing architectures, which make it difficult for on-board processing. Besides, the recent literature provides few examples for the use of HPC to implement UAV-RS generic data processing, thus more practice and attempts are required.

\subsection{A List of open-source data and algorithms}

To provide an easy starting point for researchers attempting to work on UAV-RS photogrammetric processing, we list some available resources, including tools and some algorithms. In addition, we provide a selected list of open-source UAV-RS data sets for evaluating algorithms and training deep learning models. It is noting that the open-source resource listed below is a non-exhaustive list.

\vspace{2mm}
\subsubsection{\textbf{Tools and algorithms for UAV-RS data processing}}
Some open-source tools and algorithms which can be used for UAV-RS photogrammetric processing have been proposed, shown in Tab.~\ref{Tab:Tools} and Tab.~\ref{Tab:Algorithms}. The codes of algorithms can be downloaded from respective papers. Noting that all these examples are offered with open licenses, and the corresponding papers must be acknowledged when using those codes. The rules on the respective websites apply. Please read the specific terms and conditions carefully. These available tools provide great convenience for the development of algorithms used for UAV-RS data processing, and make it easy to get started.

\begin{table}
	\renewcommand\arraystretch{1.2}
	\centering
	\caption{Some available tools for UAV-RS data processing.}
	\label{Tab:Tools}
	\begin{tabular}{m{0.28\columnwidth}<{\centering} | m{0.46\columnwidth}<{\centering}} 
		\hline
		Item                                                        & Tools                \\
		\hline
		Computer vision                                    & OpenCV and VLFeat  \\ \hline
		UAV data processing                             & OpenDroneMap (ODM) \\ \hline
				
		\multirow{2}{*}{SfM library}             & Bundler, VisualSFM, OpenMVG, \\ 
		                                                               & MVE, Theia and ColMap \\  \hline
		
		Dense matching                                      & MicMac, SURE and PMVS \\ \hline
		
		\multirow{2}{*}{Image stitching}	    & Image composition editor (ICE),\\
		                                                               & Autostitch and Photoshop \\ \hline
		
		\multirow{2}{*}{DL frameworks}		 & TensorFlow, Torch, Caffe, \\
		                                                               & Theano and MXNet \\	
		\hline
	\end{tabular}
\end{table}

\begin{table}
	\renewcommand\arraystretch{1.3}
	\centering
	\caption{Some available algorithms for UAV-RS data processing.}
	\label{Tab:Algorithms}
	\begin{tabular}{p{0.26\columnwidth}<{\centering} | p{0.66\columnwidth}<{\centering}} 
		\hline
		Item                                                        & Algorithms                \\
		\hline
		Camera calibration                                 		
		& Extended Hough transform~\cite{Bukhari2013}, 	One-parameter division model~\cite{Alemanipol2014}, MLEO~\cite{ZhangMleo2015}, CNN based~\cite{RongCalibCNN2016}    \\ \hline
		
	    Image matching	
		& TILDE~\cite{VerdieYFL2015}, TCD~\cite{zhangYu2017}, ASJ detector~\cite{XueXia2018}, Spread-out Descriptor~\cite{ZhangYKC2017}, CVM-Net~\cite{Hu2018CVMNet} \\\hline  
		
		Aerial triangulation	
		&PoseNet~\cite{Kendall2015}, SfMLearner~\cite{zhou2017unsupervised}, 1DSfM~\cite{Wilson2014}\\\hline
		
		Dense reconstruction		
		& PMVS~\cite{FurukawaPonce2010a}, MVSNet~\cite{YaoMVSNet2018}, DeepMVS~\cite{HuangDeepMVS2018} \\\hline
				
		Image stitching		
		& APAP~\cite{ZaragozaChin2014}, ELA~\cite{LiELA2018}, NISwGSP~\cite{ChenYS2016gsp}, Planar mosaicking~\cite{XiaYao2017}\\\hline
		
		Multisensor registration	
		& LGHD~\cite{ChenDEO2018}, HOPC~\cite{YeSBS2017}\\	
		\hline
	\end{tabular}
\end{table}

\vspace{2mm}
\subsubsection{\textbf{Open-source remote-sensing Data}}
Large data sets are in demand to train deep learning models with good generalization, both for fine-tuning models and for training networks from scratch. They are also useful for evaluating the performance of various algorithms. However, recent years have seen few works about open-source UAV-RS data sets made public, which requires more efforts. Some data sets are as follow.

\begin{itemize}
	\item \emph{Fisheye rectification data set}~\cite{YinFisheyenet2018}: This is a synthesized dataset that covers various scenes and distortion parameter settings for rectification of fisheye images. It contains 2,550 source images, each of which is used to generate 10 samples with various distortion parameter settings.

    \item \emph{ISPRS/EuroSDR benchmark for multi-platform photogrammetry}~\cite{NexISPRSBench2015}: 
    The ISPRS/EuroSDR provides three data sets (\emph{i.e.} oblique airborne, UAV-based and terrestrial images) over the two cities of Dortmund (Germany) and Zurich (Switzerland). These data sets are used to assess different algorithms for image orientation and dense matching. Terrestrial laser scanning, aerial laser scanning as well as topographic networks and GNSS points were acquired as ground truth to compare 3D coordinates on check points and evaluate cross sections and residuals on generated point cloud surfaces. 

    \item \emph{Urban Drone Dataset (UDD)}~\cite{ChenSFMData2018}: This data set is a collection of UAV images extracted from 10 video sequences used for structure from motion. About 1\%-2\% data (about 205 frames) are annotated by 3 semantic classes (vegetation, building and free space) for semantic constraints in 3D reconstruction. The data is acquired by DJI-Phantom 4 at altitudes between 60 and 100 m over the four cities of Beijing, Huludao, Zhengzhou and Cangzhouo (China). 

    \item \emph{UAV image mosaicking data set}~\cite{XuOu2016}: This data set consists of hundreds of images captured by the UAV. The corresponding DOMs are generated by DPGrid, which can be used as the golden standard to evaluate your mosaicking algorithms. 

\end{itemize}

\section{Applications}
\label{sec:applications}

UAV-based remote sensing has attracted increasing attentions in recent years. It is widely used to quickly acquire high-resolution data in small areas or fly on high-risk or difficult regions to carry out remote-sensing tasks. Based on remote-sensing products, \emph{e.g.} DOM, DEM and 3D models, UAV-RS is applied for urban planning, engineering monitoring, ecological research, and so on. The applications of UAV-based remote sensing seem to be unlimited and continually growing. 

Recognizing space limitations, we focus on some potential and novel applications in this section. Some other mature or long-standing applications, such as precision agriculture~\cite{ZhangJohn2012}, coastal and polar monitoring~\cite{GoncalvesHenriques2015,BhardwajSam2016,Leary2017}, disaster and emergency monitoring~\cite{GomezPurdie2016}, and atmospheric monitoring~\cite{Elston2015}, could be not discussed here and can be referred to papers~\cite{Colomina2014, Pajares2015, HazimUAVS2018,ParshinUAVMagnetic2018}. In fact, other applications not discussed here are still booming and deserve attention.

\subsection{Urban planning and management}

In recent years, the applications of UAV-based remote sensing in urban planning and management has experienced exponentially growth, including inspection of infrastructure conditions, monitoring of urban environments and transportation, 3D landscapes mapping and urban planning~\cite{LiuChen2014, HamHan2016}.

\vspace{2mm}
\subsubsection{\textbf{3D city modeling}} 
The camera-based UAV system provides a powerful tool to obtain 3D models of urban scenarios in a non-invasive and low-cost way. The city components are reconstructed for urban planning, including visualization, measurement, inspection and illegal building monitoring~\cite{BiljeckiSLZ2015APP}.

A pilot project was conducted using UAV-RS to build high-resolution urban models at large scale in complex urban areas in~\cite{QinGrun2013}. Specifically, a Falcon octocopter UAV equipped with a Sony camera was employed to acquire images under 150 \emph{m} and generate 3D models of campus with 6$\sim$8 \emph{cm} accuracy. GIS-layers and near infrared channel are also combined to help reconstruction of urban terrain as well as extraction of streets, buildings and vegetation. 

\begin{figure}[t]
	\centering
	\includegraphics[width=0.4\linewidth]{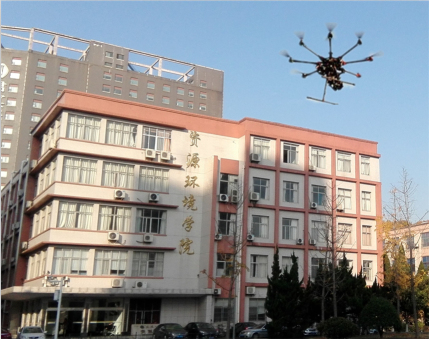}
	\includegraphics[width=0.4\linewidth]{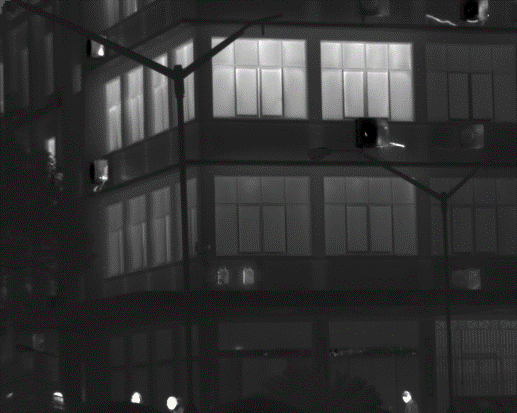} \\
	\includegraphics[width=0.8\linewidth]{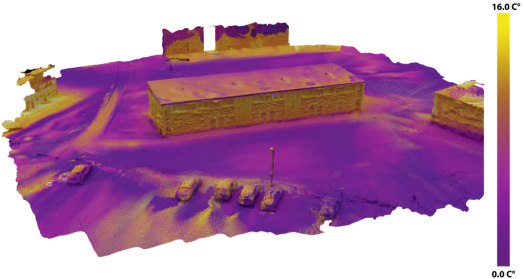}
	\caption{Monitoring of thermal information of buildings by UAVs~\cite{RakhaBuilt2018}. Top: Data acquisition for building inspection by UAVs (left) and infrared images of buildings which reflect thermal information (right). Bottom: 3D thermal model of building.}
	\label{fig:UAVBuildHeat}
\end{figure}

\vspace{2mm}
\subsubsection{\textbf{Built environment monitoring and assessment}}

UAV-RS benefits for monitoring and assessing build environment to maintain and improve our living conditions.

Regular inspection of build environment is necessary to assess health of infrastructures and identify any faults at an early stage so as to perform the required maintenance. For instance, the damage of buildings was assessed based on gaps in UAV image-derived 3D point clouds, which were identified by SVM and random forests based on the surrounding damage patterns~\cite{Vetrivel2015}. UAV visible and infrared images are acquired to monitor the condition and structural health of bridges, including bridge deterioration, deck delamination, aging of road surfaces, crack and deformation detection~\cite{EllenbergKontsos2016}. The inspection help engineers prioritize critical repair and maintenance needs.

UAV-based infrared remote sensing present an opportunity to inspect and analyze urban thermal environment, building performance and heat transfer at a micro scale so as to maintain the energy efficiency of such infrastructure and building stock~\cite{RakhaBuilt2018}. An example of monitoring thermal environment in buildings using UAVs is shown in Fig.~\ref{fig:UAVBuildHeat}. 3D thermal model of buildings are generated for monitoring and analysis of building heat distribution and leakages, to help retrofitting of aging and energy inefficient building stock and infrastructure.

Urban informal settlement are classified and identified based on very high resolution and up-to-date data from UAVs to support informal settlement upgrading projects~\cite{GevaertP2017}. Urban vegetation mapping are performed to identify land cover types and vegetation coverage in urban areas, which is significant to help planners take measures for urban ecosystem optimization and climate improvement~\cite{FengLiu2015}.

\begin{table*}[!htb]
	\renewcommand\arraystretch{1.2}  
	\centering
	\caption{Researches on UAV-based traffic target detection and tracking.}
	\label{Tab:Traffic}
	\begin{tabular}{m{0.04\linewidth} | m{0.23\linewidth} | m{0.30\linewidth} | m{0.25\linewidth}}
		\hline
		Ref.      &  Platforms       & Aim of study     &  Methods   \\ 
		\hline
		
		\cite{GonzaloStephen2012}  
		& Rotary-wing UAV, RGB camera         & Detect and track road moving objects     & Optical flow \\  
		
		\cite{PerUmut2012}   
		& UAV, gimballed vision sensor            & Road bounded vehicles search and tracking     & Particle filter, point-mass filter \\  
		
		\cite{MoranduzzoM2014aut}   
		&Rotary-wing UAV, RGB camera         & Car detection and counting     & SIFT+SVM    \\  
		
		\cite{MoranduzzoM2014det}      
		& Rotary-wing UAV, RGB camera        & Car detection, including the number, position and orientation of cars     & Similarity measure  \\  
		
		\cite{LiuMattyus2015}   
		& UAV, RGB camera                            & Vehicle detection  & Multiclass classifier  \\   
		
		\cite{XuYuWWM2016}  
		& Rotary-wing UAV, Gopro camera      & Vehicle detection      & Viola-Jones and HOG+SVM  \\  
		
		\cite{FuDuanKK2016}   
		& Rotary-wing UAV, RGB cameras      & Track container, moving car and people  & Optical flow  \\ 
		
		\cite{MaWuYu2016}     
		& Rotary-wing UAV, infrared camera   & Pedestrian detection and tracking   & Classification, optical flow    \\  
		
		\cite{HsieDrone2017}   
		& Rotary-wing UAV, RGB camera      & Detect, count and localize cars  & Deep CNN   \\   
		
		\cite{LiYeungTrack2017}  
		& Rotary-wing UAV, RGB camera     & Visual object tracking (\emph{e.g.} people and cars)   & Deep CNN   \\   
		
		\cite{YangDetection2018}  
		& UAV, visible camera   & Vehicle detection  & Deep CNN   \\   
		
		\cite{DuUAVBench2018}  
		& Rotary-wing UAV, RGB camera  & A large dataset for object detection and tracking  & Deep CNN  \\  
		\hline
	\end{tabular}
\end{table*}

\vspace{2mm}
\subsubsection{\textbf{Urban traffic monitoring}}

UAVs, like eyes on the sky, provide the ``above-the-head'' point of view for surveillance, especially in traffic monitoring~\cite{LeitloffRosenbaum2014, Barmpounakis2016UAVTransp}, including detection and tracking of traffic targets, crowd monitoring, estimation of density, capacity and traffic flow. Traffic monitoring is beneficial to ensure security, optimize urban mobility, avoid traffic jams and congestions, analyze and solve environmental problems affecting urban areas. 

Traffic target detection and tracking are two essential technologies in urban traffic monitoring.  However, UAV-based detection and tracking is a challenging task, owing to object appearance changes caused by different situations, such as occlusion, shape deformation, large pose variation, onboard mechanical vibration and various ambient illumination~\cite{FuDuanKK2016}. Numerous methods are proposed focusing on UAV-based traffic target detection and tracking, shown in Tab.~\ref{Tab:Traffic}.

\begin{figure}[t]
	\centering	
	\includegraphics[height=0.4\linewidth]{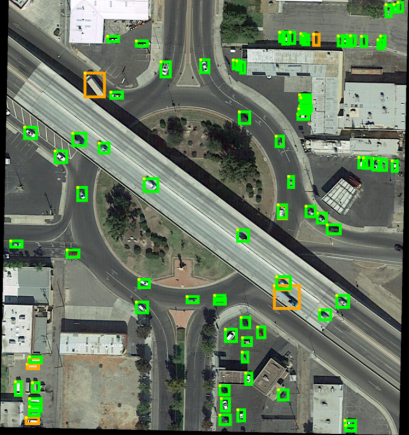}	
	\includegraphics[height=0.4\linewidth]{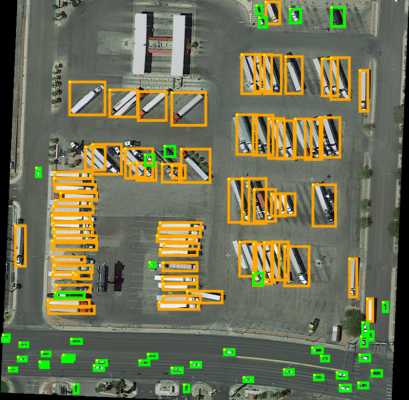}		
	\caption{Vehicle detection and traffic monitoring by UAVs based on deep learning. Left: Vehicle detection in crossing. Right: Vehicle detection in road and park. Orange boxes denote large cars and green boxes denote small cars.}
	\label{fig:Surveillance}
\end{figure}

Various traffic targets, including cars, pedestrian, roads and bridges, are detected, localized and tracked based on UAV visible or infrared cameras. An example of vehicle detection and traffic monitoring can be seen in Fig.~\ref{fig:Surveillance}. Except for traffic monitoring, UAV-RS can also be used for traffic emergency monitoring and document, pedestrian-vehicle crash analysis and pedestrian/vehicle behavior study. In~\cite{highDdataset2018}, the camera-equipped UAVs are used to record road traffic data, measure every vehicle’s position and movements from an aerial perspective for analyzing naturalistic vehicle trajectory and naturalistic driving behavior.

\subsection{Engineering monitoring}

UAVs provide a bird's-eye view solution for engineers to plan, build and maintain projects~\cite{LiuChen2014}. With UAVs, construction managers can monitor the entire site with better visibility, so that they are more informed about project progress. In addition, engineering observation and inspection by UAVs can ensure field staff safety, reduce production risks and increase on-site productivity when compared with artificial means. Recently, UAV-based remote sensing are widely applied in oil and gas pipelines monitoring, power infrastructure monitoring, mine areas monitoring, civil engineering, engineering deformation monitoring and railway monitoring~\cite{SiebertTeizer2014}.   

\vspace{2mm}
\subsubsection{\textbf{Oil and gas pipeline monitoring}}

UAV provides a cost-effective solution for monitoring oil/gas pipelines and its surroundings~\cite{BarchynHMB2017}, in contrast  to conventional foot patrols and aerial surveillance by small planes or helicopters which are time-consuming and high-cost. UAVs are used to map pipelines and the surroundings, detect leakage and theft, monitor soil movement and prevent third-party interference, etc~\cite{GomezGreen2017}. Generally, frequent observation by UAVs help timely identify corrosion and damage along pipelines so as to make proactive responses and maintenance. For identification of hydrocarbon leak, thermal infrared sensors are widely used to detect the temperature differences between the soil and fluids (\emph{i.e.} hydrocarbons). For detection of gas emission and leak, gas detection sensors are applied. Although gas may diffuse or disperse into atmosphere, especially in windy weather, the leakage location can be estimated by the concentration of gas.	
	
\vspace{2mm}
\subsubsection{\textbf{Power infrastructure monitoring}}

\begin{figure}[tbp]
	\centering
	\subfigure[]{
		\label{fig:Powermonitor.1}	
		\includegraphics[height=0.2\linewidth]{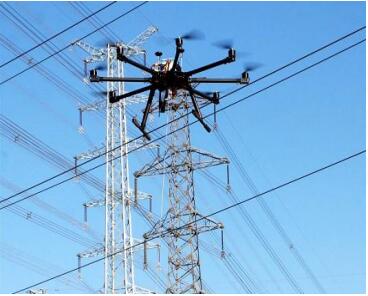} }
	\subfigure[]{
		\label{fig:Powermonitor.2}	
		\includegraphics[height=0.2\linewidth]{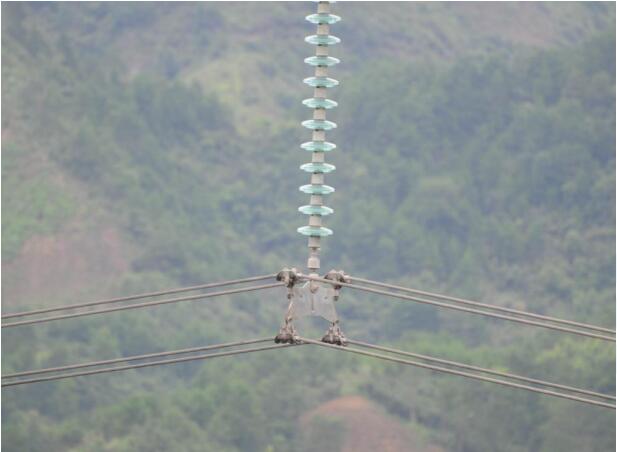} }
	\subfigure[]{
		\label{fig:Powermonitor.3}	
		\includegraphics[height=0.2\linewidth]{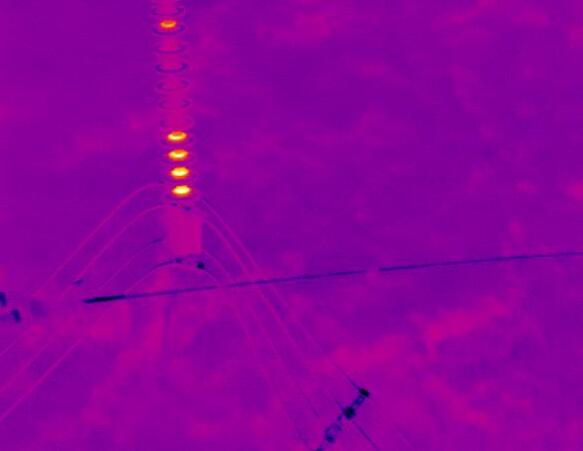} } \\ [-1mm]
	\subfigure[]{
		\label{fig:Powermonitor.4}	
		\includegraphics[width=0.87\linewidth]{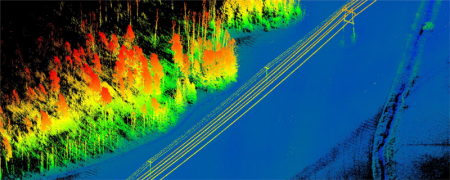} }\\
	\caption{An example of power facilities monitoring. (a) UAV-based power inspection. (b) Visible image of insulator. (c) Infrared image of heating insulator. (d) Laser scanner data of power line corridor acquired by UAVs. (a)-(c) are provided by Xinqiao Wu, and (d) is from Leena~\emph{et al.}~\cite{LeenaMatti2016}.}
	\label{fig:Powermonitor}
\end{figure}

UAV-RS have been also widely applied to monitor power infrastructures, including power lines, poles, pylons and power station, during the period of plan, construction and maintenance of electric grids~\cite{LeenaMatti2016}. An example of power facilities monitoring is shown in Fig.~\ref{fig:Powermonitor}. 

In fact, it is an important but challenging task to detect power facilities from cluttered background and identify their defects~\cite{Lim2018MultiUAV}. As one of the most important power infrastructures, power lines are often detected by line-based detection, supervised classification or 3D point cloud-based methods~\cite{ZhuJuha2014}. Other power equipments are also detected, including conductors, insulators (glass/porcelain cap-and-pin and composite insulator), tower bodies, spacers, dampers, clamps, arcing horns and vegetation in corridors. The defects of power facilities (\emph{e.g.} mechanical damage and corrosion) and the distance between vegetation/buildings and power lines are often identified based on visual inspection, thermography and ultraviolet cameras~\cite{JakaFB2010}.

Besides, the radioactivity of nuclear plant was assessed using radiation sensors-equipped UAVs, including mapping evolving distribution of radiation, analyzing the contributing radionuclide species and the radiological or chemo-toxicity risks~\cite{MartinMoore2016}. The influence of power plant on the surrounding environment is also monitored, which uses UAVs with infrared cameras to map temperature profiles of thermal effluent at a coal burning power plant in~\cite{DemarioLP2017}.

\vspace{2mm}
\subsubsection{\textbf{Mine areas monitoring}}

Mine areas are usually large and located in distant mountainous areas, which bring about challenges for monitoring by traditional methods. UAV-RS offers a promising way for mapping, monitoring and assessment of mine areas and their surroundings.

UAV-RS are often used to monitor mining activities and geomorphic changes of mining area, which can provide guidance for mine production and safety. For instance, surface moisture of peat production area is monitored to ensure environmental safety of peat production using UAVs with hyperspectral frame cameras~\cite{EijaMatti2016}. Side slopes are mapped for mine area inventory and change monitoring based on terrestrial laser scanning and UAV photogrammetry~\cite{TongLiu2015}. Orthophotos and 3D models of mine areas are generated to assess the detailed structural-geological setting and identify potentially unstable zones so as to evaluate safety conditions and plan for proper remediation. 

Besides, dust emission of mine tailings has a big influence on surrounding environment of mine areas, which can be mitigated by monitoring and controlling moisture of mine tailings. In~\cite{ZwisslerOV2017}, thermal sensors are mounted on UAVs to acquire data of iron mine tailings to map the spatial and temporal variations in moisture content of surface tailings. The relationship between moisture and strength of mine tailings is analyzed to help management of mine tailings.

\subsection{Ecological and environmental monitoring}

For ecological and environmental research, most areas are too remote or too dangerous to be thoroughly surveyed. Besides, most ecological experiments that involve many repetitive tasks are difficult to be conducted due to lack of necessary manpower and time or high cost of manned aerial survey. The emerging of UAVs opens new opportunities and revolutionizes the acquisition of ecological and environmental data~\cite{LianSerge2012}. Moreover, UAVs make it possible to monitor ecological phenomena at appropriate spatial and temporal resolutions, even individual organisms and their spatio-temporal dynamics at close range~\cite{AndersonGaston2013}. Recent years have seen rapid expansion of UAV-RS in ecological and environmental research, monitoring, management and conservation.

\vspace{2mm}
\subsubsection{\textbf{Population ecology}}

Population ecology aims to study, monitor and manage wildlife and their habitats. It is challenging for ecologists to approach sensitive or aggressive species and access remote habitats. UAV-RS makes regular wildlife monitoring, management and protection possible and provides more precise results compared with traditional ground-based surveying~\cite{HodgsonBM2016}. It is often applied to estimate populations/abundance and distribution, monitor wildlife behavior, map habitat and range, perform wildlife conservation including anti-poaching and illegal trade surveillance~\cite{JulieJJ2015}, shown in Tab.~\ref{Tab:Population}.

\begin{table*}[!htb]
	\renewcommand\arraystretch{1.2}  
	\centering
	\caption{Researches on population ecology using UAV-RS.}
	\label{Tab:Population}
	\begin{tabular}{m{0.18\linewidth} | m{0.4\linewidth} | m{0.3\linewidth}}
		\hline
		Item                       & Contents                 & Methods \\ 
		\hline
		Population estimation    
		& Wildlife identification, enumeration, and estimation of their population status, \emph{e.g.} amount, abundance and distribution
		& Manual visual inspection~\cite{HodgsonKelly2013}, deformable part-based mode~\cite{JanCamiel2014}, threshold and template matching~\cite{GonzalezMontes2016}, classification~\cite{SeymourDale2017} \\ \hline   
	
		Wildlife tracking
		& Explore animal behaviors (\emph{e.g.} migratory patterns) and habitats so as to sustain species and prevent extinction
		& Long-term target tracking, acoustic biotelemetry,  radio collar tracking~\cite{KornerSpeck2010} \\ \hline
	    
	    Habitat and range mapping
	    & Monitor habitat status, including vegetation distribution and coverage, seasonal or environmental changes of habitats    
	    & Orthophoto generation, classification~\cite{FlynnChapra2014} \\ \hline
	  
	    Conservation of wildlife
	    & Anti-poaching surveillance and wildlife protection, \emph{e.g.} detecting animals, people/boats acting as poachers, and illegal activities
	    & Target detection~\cite{MargaritaRoel2014} \\     	
		\hline
	\end{tabular}
\end{table*}  

Most of species that have been monitored by UAVs contains large terrestrial mammals (\emph{e.g.} elephants), aquatic mammals (\emph{e.g.} whales) and birds (\emph{e.g.} snow geese). However, it is noting that UAVs may disturb wildlife and thus cause behavioral and physiological responses when flying at low altitude and high speed for close observation. With the increasing use of UAVs, particularly in research of vulnerable or sensitive species, there is a need to balance the potential disturbance to the animals with benefits obtained from UAV-based observation~\cite{ChristieGB2016}.

\vspace{2mm}
\subsubsection{\textbf{Natural resources monitoring}}
	
Natural resources, \emph{e.g.} forest, grassland, soil and water, are of great need for monitoring, management and conservation, which gain increasing benefits from UAV-RS recently~\cite{ShahbaziTM2014}. Here we take forest and grassland as examples to illustrate applications of UAV-RS.

\begin{table*}[!htb]
	\renewcommand\arraystretch{1.2}  
	\centering
	\caption{Researches on forest monitoring using UAV-RS.}
	\label{Tab:Natural}
	\begin{tabular}{m{0.08\linewidth}<{\centering} | m{0.38\linewidth} | m{0.5\linewidth} }
		\hline
		Item                       & Contents              & Methods \\
		\hline
		\multirow{1}{*}{\shortstack{Forest\\ structure} }  
		& - Forest 3D structural characterization, including DTM, canopy height model and canopy surface model; 
		& - 3D structures: SfM photogrammetry, LiDAR and profiling radar~\cite{ChenHakala2017}; \\ \hline

		\multirow{2}{*}{\shortstack{Forest\\ inventory}}		
		& - Measure properties about geometry structure and spatial distribution of trees; 
		& - Plot-level metrics: canopy points or image classification~\cite{WallaceLW2014eval};  \\
		
		& - Estimate terrain/under-story height, plot-/tree-level metrics. 
		& - Tree-level metrics: canopy height model~\cite{WallaceMusk2014}.  \\ \hline

		\multirow{2}{*}{\shortstack{Forest\\ biomass}} 
		& Above-ground biomass estimation 
		& - UAV-based L-band radar~\cite{ZhangNi2017}; \\
		
		& 
		& - Vertical information + L-band radar~\cite{SaatchiMarlier2011}. \\ \hline
		 
		 \multirow{2}{*}{\shortstack{Forest\\ biodiversity}}  
		 & Monitor forest biodiversity at spatial and temporal scale	 
		 & - Quantification of canopy spatial structures and gap patterns~\cite{GetzinNuske2014}; \\
		 
		 &  
		 & - Fallen trees detection and their spatio-temporal variation analysis~\cite{BunnellHoude2010}. \\ \hline
		 
		  \multirow{1}{*}{\shortstack{Forest health\\ monitoring}} 		 
		 & Monitor forest health, \emph{e.g.} identification of disease and insect pest damage 		 
		 & Multi- and hyper-spectral remote sensing,  dense point clouds~\cite{RoopeEija2015,BrovkinaUAVForest2018}   \\ \hline
		 
		  \multirow{3}{*}{\shortstack{Forest fire\\ monitoring}} 	 
		 & 
		 & - Before fires: forest prevention, \emph{e.g.} create fire risk maps, (3D) vegetation maps;\\
		 
		 & Forest fire monitoring, detection and fighting
		 & - During fires: detect active fires, locate fires, predict fire propagation; \\ 
		 
		 &
		 & - After fires: detect active embers, map burned areas and assess fire effects~\cite{YuanZhang2015} \\


		\hline
	\end{tabular}
\end{table*}

\begin{itemize}
	\item[-] \emph{Forest monitoring.} Forest resources are the most common scenarios in UAV applications~\cite{TorresanBerton2017}, including forest structure estimation, forest inventory, biomass estimation, biodiversity, disease and pests detection, and forest fire monitoring, shown in Tab.~\ref{Tab:Natural}. UAV-RS takes a strong advantage in small-area forest monitoring. The continued explosion of forest monitoring applications relies mainly on the flight endurance and observation capability of payload.
	
	\item[-] \emph{Grassland and shrubland monitoring.} Grassland or shrubland are often located in remote areas with low population density, which poses challenges for their assessment, monitoring and management. Due to flexibility, high resolution and low cost, UAV-RS holds great potential in grassland and shrubland monitoring. Some examples are shown in Tab.~\ref{Tab:Grassland}. UAV-RS is an emerging technology that has gained growing popularity in grassland monitoring. However, the use of high-standard multi- or hyper-spectral sensors, which are beneficial for species classification, remains a challenge due to the high weight. Besides, it is also encouraged to explore the optimal spatial resolution for studying different vegetation characteristics.
		
\end{itemize}

\begin{table*}[!htb]
	\renewcommand\arraystretch{1.2}  
	\centering
	\caption{Researches on UAV-based grassland and shrubland monitoring.}
	\label{Tab:Grassland}
	\begin{tabular}{m{0.04\linewidth} | m{0.14\linewidth} | m{0.35\linewidth} | m{0.38\linewidth}}
		\hline
		Ref.      &  Platforms   & Payloads    & Aim of study        \\ 
		\hline
		
		\cite{AndreaAlbert2009}   & Fixed-wing UAV    & Canon SD 550
		& Differentiate bare ground, shrubs, and herbaceous vegetation in an arid rangeland \\
				
		\cite{AndreaMark2011}      & Fixed-wing UAV     & Color video camera, Canon SD 900, Mini MCA-6
		& Rangeland  species-level vegetation classification \\
                 
         \cite{CapolupoKooistra2015}   & Octocopter UAV   & Panasonic GX1 digital camera, hyperspectral camera
         & Estimate plant traits of grasslands and monitor grassland health status     \\
         
         \cite{BuerenBurkart2015}    & Rotary-wing UAV    & RGB camera, near-infrared camera, MCA6 and hyperspectral camera
         & Evaluate the applicability of four optical cameras for grassland monitoring  \\
         
         \cite{ChenYi2016}      & Quadcopter UAV     & GoPro Hero digital camera
         & Estimation of  fractional vegetation cover of  alpine grassland \\
         
         \cite{LopatinGrassland2017}   & Simulation platform  & AISA + Eagle imaging spectrometer 
         & Hyperspectral classification of grassland species at the level of individuals\\
         
         \cite{Schmidt2017}  & UAV   & RGB camera, hyperspectral camera
         & Mapping the conservation status of Calluna-dominated Natura 2000 dwarf shrub habitats \\

		\hline
	\end{tabular}
\end{table*}

\vspace{2mm}
\subsubsection{\textbf{Species distribution modeling}} 

Over the past decades, a considerable amount of work has been performed to map species distributions and use these collected information to identify suitable habitats. Species distribution modeling is one such work, which models species geographic distributions based on correlations between known occurrence records and the environmental conditions at occurrence localities~\cite{GomesSDM2018}. It has been widely applied in selecting nature reserves, predicting the effects of environmental change on species range and assessing the risk of species invasions~\cite{Mullerova2017}. 

Due to the spatial biases and insufficient sampling of conventional field surveys, UAV-RS has become a very effective technology to supply species occurrence data recently, attributable to its ability to quickly and repeatedly acquire very high-spatial resolution imagery with low cost~\cite{HeSDM2015}. For instance, UAV-RS is used to detect plant/animal species in terrestrial and aquatic ecosystems, estimate their populations and distribution patterns, and identify important habitat (\emph{e.g.} stopovers on migratory routes, breeding grounds)~\cite{HodgsonKelly2013,FlynnChapra2014,SeymourDale2017}. Moreover, UAV-RS provides a timely and on-demand data acquisition, which allows a more dynamic manner to understand habitat suitability and species range expansion or contraction.

However, UAV-RS  may also cause uncertainty and errors for species distribution modeling. These errors mainly come from data acquisition and processing algorithms, such as species classification. Thus, strict data acquisition and high-precision data processing and analysis are necessary.

\vspace{2mm}
\subsubsection{\textbf{Environmental monitoring and conservation}}
UAVs are used to monitor environmental process and changes at spatial and temporal scales, which is challenging by conventional remote-sensing platforms~\cite{PerryRyan2011}, \emph{e.g.} mudflat evolution and morphological dynamics~\cite{JaudGrasso2016}. 
Besides, environmental pollution monitoring greatly benefits from UAV-based remote sensing. UAVs equipped with multi-spectral sensors are employed to map trophic state of reservoir and investigate water pollution for water quality monitoring in~\cite{SuChou2015}. 
Soil erosion, degradation and pollution are also monitored based on UAV DTMs and orthophotos. For instance, soil copper contamination was detected based on hydrological models using a multi-rotator UAV, and copper accumulation points were estimated at plot scales based on micro-rill network modeling and wetland prediction indexes~\cite{CapolupoPindozzi2015}.

\begin{figure}[t]
	\centering	
	\includegraphics[width=0.7\linewidth]{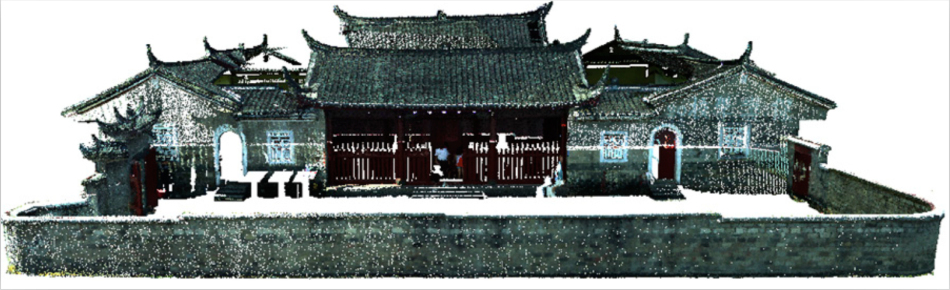}\\	
	\includegraphics[width=0.7\linewidth]{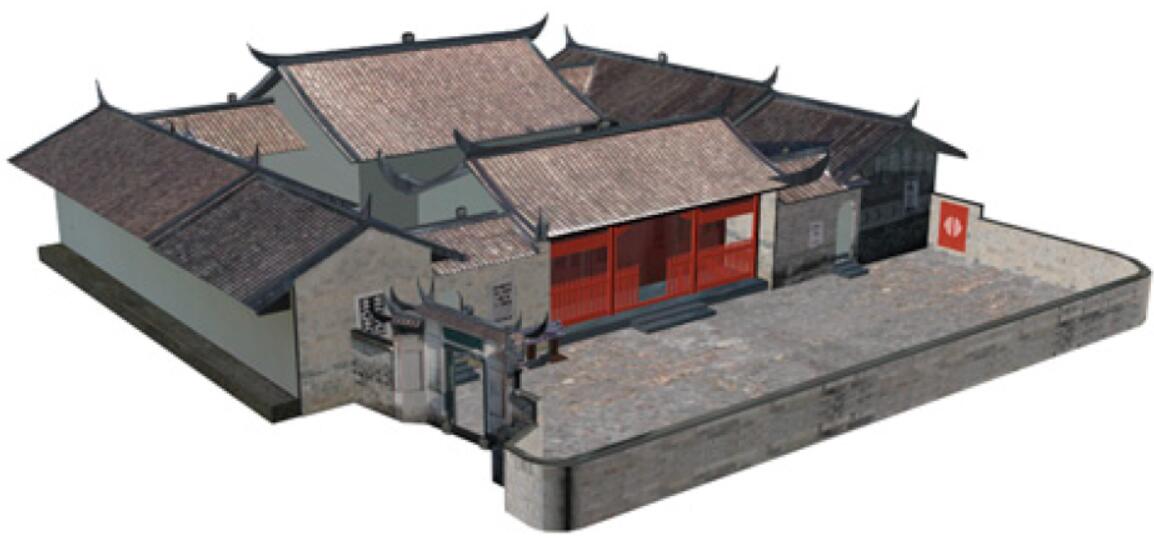}		
	\caption{3D digitalization of cultural heritage for recording and conservation~\cite{XuWuShen2014}. (a) Dense point cloud of Gutian conference monument. (b) Photo-realistic 3D model of the monument.}
	\label{fig:Culture}
\end{figure}

\subsection{Archeology and cultural heritage}

Archeology and cultural heritage is a promising application for UAV-based remote sensing~\cite{Fernandez2015image}. 

UAVs are generally adopted to conduct photogrammetric survey and mapping, documentation and preservation of archaeological site~\cite{FranciscoMaria2016}. In addition, it is also used for archaeological detection and discovery. In archeology, buried features may produce small changes or anomalies in surface conditions, which can be detected and measured based on UAVs with spectroradiometer, digital or thermal cameras~\cite{LinNovo2011}.

For cultural heritage, UAVs are often employed to produce high-quality 3D recordings and presentations for documentation, inspection, conservation, restoration and museum exhibitions~\cite{JoseMarcello2016}.
Multiple platforms,~\emph{e.g.} terrestrial laser scanner, ultralight aerial platform, unmanned aerial vehicle and terrestrial photogrammetry, are often integrated to acquire multi-view data for 3D reconstruction and visualization of cultural relics. In Fig.~\ref{fig:Culture}, a camera-equipped UAV is integrated with a terrestrial laser scanner to facilitate complete data acquisition of historical site, which building facades are captured by terrestrial laser scanner and building roofs are acquired by UAV photogrammetry~\cite{XuWuShen2014}.  

Restoration of heritage are usually based on precision 3D data. In~\cite{ChenHu2016}, a virtual restoration approach was proposed for the ancient plank road. The UAV and terrestrial laser scanner were used to collect detailed 3D data of existing plank roads, which were processed to determine the forms of plank roads and restore each component with detailed sizes based on mechanical analysis. The virtual restoration model was then generated by adding components and background scene into 3D model of plank roads.

\subsection{Human and social understanding}

The aerial view of UAV-RS makes it a potential solution to help describe, model, predict and understanding human behaviors and interaction with society. 

\begin{figure}[t]
	\centering	
	\includegraphics[height=0.3\linewidth]{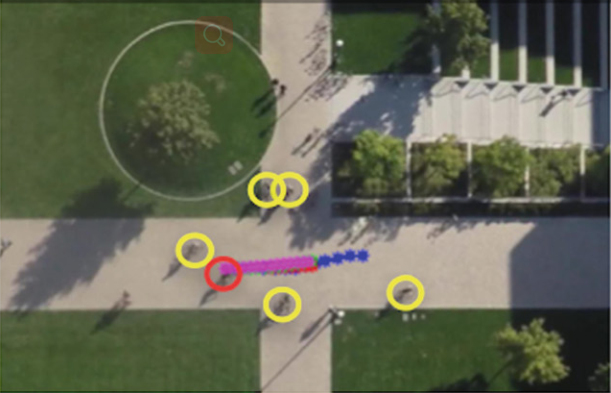}
	\includegraphics[height=0.3\linewidth]{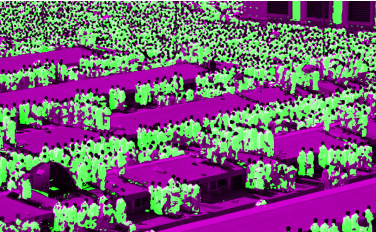}	
	\caption{Left: Pedestrian trajectory prediction~\cite{Robicquet2016}. Right: Crowd monitoring~\cite{ShearyCrowd2017}.}
	\label{fig:Hum}
\end{figure}

In~\cite{Robicquet2016}, UAVs are used to collect videos of various types of targets, \emph{e.g.} pedestrians, bikers, cars and buses, to understand pedestrian trajectory and their interact with the physical space as well as with the targets that populate such spaces. This could make a great contribution to pedestrian tracking, target trajectory prediction and activity understanding~\cite{BarekatainUAVAction17}. 
In~\cite{highDdataset2018}, researchers adopt a camera-equipped UAV to record naturalistic vehicle trajectory and naturalistic behavior of road users, which is intended for scenario-based safety validation of highly automated vehicles. The data can also be used to contribute on driver models and road user prediction models. 
Besides, UAV-RS is beneficial for crowd risk analysis and crowd safety, especially in mass gatherings of people related to sports, religious and cultural activities~\cite{ShearyCrowd2017, kang2018counting}. UAV-RS flexibly provides high-resolution and real-time on-the-spot data for people detection, crowd density estimation and crowd behavior analysis so as to make effectively response to potential risk situation. Fig.~\ref{fig:Hum} shows some examples. 

Recent studies provide only a few works about human and social understanding using UAV-RS. However, with the popularity of UAVs available to everyone, we can see a huge rising research topic in UAV-RS.

\section{Prospectives}
\label{sec:prospect}

Thanks to the progress of UAV platforms and small-size remote-sensing sensors, as well as the improvement of UAV regulations and the opening of market, UAV-RS is gaining a lot of popularity in the remote-sensing community. However, a lot of challenges remain which require more efforts.

\begin{itemize}
	\item[-] \emph{UAV platforms.} Due to the light weight and small size, UAVs often suffer from some inherent defects, including platform instability, limited payload capacity and short flight endurance, which pose challenges for acquisition of reliable remote-sensing data and high-precision data processing.
	
	\item[-] \emph{Remote-sensing sensors.} Weight and energy consumption are the main limitation for remote-sensing sensors. Thus, it is difficult to use high-precise navigation system, high-standard multi-/hyper-spectral camera, LiDAR, radar, and even massively parallel platforms for onboard processing in small UAVs.
	
	\item[-] \emph{UAV policy and regulations}. It is one of the major factors impeding the use of UAVs in remote-sensing community~\cite{WattsAmbrosia2012,ArthurRegula2017,StockerRegu2017}. Restrictions in the use of airspace prevent researchers from testing all possibilities. Indeed, UAVs used in civil applications have been developing faster than the corresponding legislation. The adaptations to the relevant legislation will be necessary in future. Undoubtedly, effective UAV regulations will facilitate the wider use of UAVs in remote-sensing community.
	
	\item[-] \emph{Data processing.} Some challenges have been discussed in each section of key technologies. Some other issues, such as robust, high-efficiency, automation and intelligence for data processing, are worth more efforts. Besides, how to handle massive multi-source/heterogeneous remote-sensing data is also worth considering. 

\end{itemize}

The current research trends and future insights are discussed below.

\subsection{Platforms} 

The continued trend of increasingly miniaturized components of UAV-RS promises an era of tailored systems for on-demand remote sensing at extraordinary levels of sensor precision and navigational accuracy~\cite{WattsAmbrosia2012}. 

\begin{itemize}
	\item[-] \emph{Long flight endurance} is expected for efficient remote-sensing data acquisition. Research is ongoing to improve the battery technology, including a power tethered UAV~\cite{SindhujaLav2015}, solar-powered UAV~\cite{Sun2018ResourceAF}, and beamed laser power UAV~\cite{Ouyang2018power}. Laser power beaming would enable unlimited flight endurance and in flight recharging of UAVs. Thus, such UAVs could fly day and night for weeks or possibly months without landing.
	
	\item[-] \emph{Light-weight, small-sized and high-precision remote-sensing sensors} are ongoing trend, which have been not yet sufficiently miniaturized~\cite{DengUAVSAR2018}. Continuing advances in the miniaturization of remote sensing sensors and positioning hardware is placing increasingly powerful monitoring and mapping equipment on ever smaller UAV platforms. Besides, more miniaturized sensors will be developed for UAV-RS, such as $CH_4$ detector and atmospheric sensor. Moreover, this makes multi-sensor integration easy to implement, strengthening the earth observation performance of UAV-RS.
		
	\item[-] \emph{Safe, reliable and stable UAV remote sensing systems}. Due to light weight and small size, UAV-RS often suffer from instability when there is airflow. Stable unmanned aircraft deserves more efforts~\cite{YangLin2016}. Video stabilization could be integrated into data acquisition systems~\cite{ShabayekD2012}. In addition, safe operation has become a global concern. Obstacle avoidance are often achieved based on ultrasound sensors or depth cameras, which are short-distance oriented. Deep learning-based vision may be a good support. Dynamic vision sensor, \emph{e.g.} event camera, is another promising solution. Besides, safe landing has been largely un-addressed. Deep networks can be used to learning  to estimate depth and safe landing areas for UAVs~\cite{MarcuSafeUAV2018}.
	
	\item[-] \emph{Autonomous navigation and intelligent UAVs.}  Although the fact that UAV can fly autonomously,  there remain challenges under challenging environments, such as indoor fire scene where GPS may fail. Besides, nowadays it is still required the presence of a pilot. One of the main reasons is the lack of device intelligence. This issue could be solved mainly by artificial intelligence, which is able to provide autonomous decision support and reaction to events including law awareness~\cite{LuisUAVAG2017}. For instance, deep learning can be used to learn to control UAVs and teach them to fly in complex environments~\cite{Giusti2016, Muller2018Teachfly}. We envision that UAV-RS is capable of providing the entire automated process from taking off the vehicle to processing the data and turning on the pro-active actions. To this end, more issues need to be considered, including intelligent perception of environments,  precision control, indoor/outdoor seamless navigation and positioning~\cite{LuXXZ2018,JulienUAVNavi2018}.
	
\end{itemize}

\subsection{Data processing}

The existing data processing can satisfy the majority of applications of UAVs in remote-sensing community, however, efforts remain in need to facilitate data processing more automatic, efficient and intelligent, which may improve the earth observation performance of UAV-based remote sensing.

\begin{figure}[t]
	\centering	
	\includegraphics[width=0.65\linewidth]{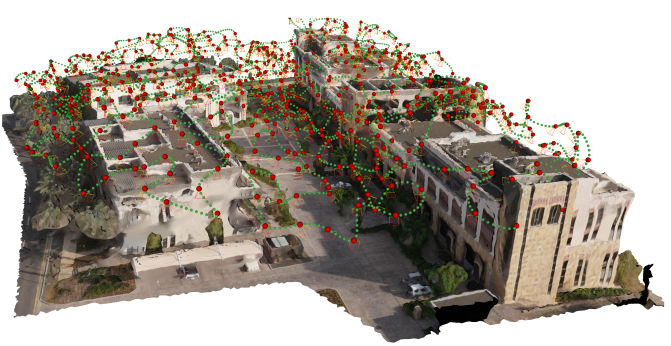}	
	\caption{Aerial path planning in urban building scenes~\cite{SmithAerialPath2018}.}
	\label{fig:Path}
\end{figure}

\begin{itemize}
	\item[-] \emph{Aerial view and path planning.} How to perform view and path planning to ensure complete and accurate coverage of the surveyed area with minimum flight time is a crucial but challenging issue. UAV-RS often acquire data either under manual control or using pre-designed flight paths, with the camera setting in a fixed direction,  \emph{e.g.} vertical or oblique. It is challenging to perform complete and dense coverage, especially in urban environment. One promising solution is to take the initial scene reconstruction from the nadir acquisition as a reference to continuously optimize the view and position~\cite{SmithAerialPath2018}. An example of aerial view and path planning is shown in Fig.~\ref{fig:Path}.
	
	\item[-] \emph{Robust data processing.} UAV-RS is expected to process remote-sensing data with different source, quality, resolution, scale, distortion, etc. which is an imperative but challenging issue. For instance, handling water covered image, cloud shelter image, arbitrary-attitude image, photography loopholes, and multi-source images (close-range, low-altitude and oblique images, or infrared and visible images) for aerial triangulation. Progress will make in the future.	
	
	\item[-] \emph{Real-time/on-board data processing.} Real-time or on-board data processing plays a significant role in UAV-RS, especially in time-critical remote sensing~\cite{KumarPatil2017}. In the wave of sensor miniaturization, FPGA and GPU are expected to be designed light-weight and low energy consumption, which are adaptive to miniaturized UAVs for on-board processing. Besides, the collected data should be processed based on high-performance computing, such as cloud computing.
	
	\item[-] \emph{Deep learning for UAV-RS.} Great success has been achieved in image classification and target detection~\cite{ZhuTM2017, UAVid2018, DOTA2018, RoI2019}, however, there is a lot of room for deep learning applied in UAV-RS 3D geometric vision, especially in image matching and pose estimation. Some critical issues should be taken into consider, including the lack of large-scale annotation data set, weakly supervised learning for limited annotated data, transfer learning for off-the-shelf deep models.

   \item[-] \emph{3D semantic computing.} There is a trend that learning to estimate 3D geometry and semantics jointly. More geometric priors should be introduced to capture the complex semantic and geometric dependencies of 3D world. Another issue is the high memory consumption, caused by the necessity to store indicator variables for every semantic label and transition, which should be considered~\cite{CherabierSemantic3D2016}.
   
   \item[-] \emph{Information mining from UAV-RS big data.} Data collected from UAV flights can reach hundreds of megabytes per hectare of surveyed area. Besides, UAVs can form a remote-sensing network to provide fast, cloudless, centimeter-level and hour-level data collection and accurate service on the Internet. This will inevitably generate massive amounts of remote sensing data. Knowledge mining from massive and heterogeneous remote-sensing data is a great challenge. Deep learning and cloud computing shed light on this issue. Besides, how to optimize data acquisition to ensure complete and accurate coverage with minimum data volume and redundancy is also crucial to reduce the difficulty of information mining.
\end{itemize}

\subsection{Applications}

With the advance of UAV platforms and remote-sensing sensors, there is potential for wider applications. The attention may shift from monitoring earth environment to human and social understanding, such as individual/group behavior analysis and infectious disease mapping~\cite{FornaceDrakeley2014}. UAV-RS also hold potential in autonomous driving community. They are adopted to extend the perception capabilities of a vehicle by using a small quadrotor to autonomously locate and observe regions occluded to the vehicle and detect potentially unsafe obstacles such as pedestrians or other cars~\cite{WallarDriving2018}. More applications are on the way.

\section{Conclusions} 
\label{sec:conclusions}

Compared to conventional platforms (\emph{e.g.} manned aircrafts and satellites), UAV-RS present several advantages: flexibility, maneuverability, efficiency, high-spatial/temporal resolution, low altitude, low cost, etc. In this article, we have systematically reviewed the current status of UAVs in remote-sensing community, including UAV-based data processing, applications, current trends and future prospectives. Some conclusions can be obtained from this survey.

\begin{itemize}
\item The inspiring advance of UAV platforms and miniaturized sensors has made UAV-RS meet the critical spatial, spectral and temporal resolution requirements, offering a powerful supplement to other remote-sensing platforms. UAV-RS holds great advantages in accommodating the ever-increasing demands for small-area, timely and fine surveying and mapping.

\item Due to the characteristics of UAV platforms, many specialized data-processing technologies are designed for UAV-RS. Technologically speaking, UAV-RS are mature enough to support the development of generic geo-information products and services. With the progress of artificial intelligence (\emph{e.g.} deep learning) and robotics, UAV-RS will experience a tremendous technological leap towards automatic, efficient and intelligent services.  
 
\item Many current UAV-RS data-processing software is commercially available, which promotes UAV-RS flourish in remote-sensing applications. With the development of UAV-RS, the applications of UAV-based remote sensing will be continually growing. 

\end{itemize}

Noting that challenges still exist and hinder the progress of UAV-RS. Numerous research is required, which is being performed with the advantage of low entrance barriers. Rapid advancement of UAV-RS seems to be unstoppable and more new technologies and applications in UAV-RS will be seen definitely in coming years.


%

\section{Author information}

\begin{IEEEbiographynophoto}{\textbf{Tian-Zhu Xiang}}
	(tzxiang@whu.edu.cn) received the B.S. and M.S. degree in photogrammetry and remote sensing from Wuhan University, Wuhan, China, in 2012 and 2015, respectively. He is currently working toward the Ph.D. degree at the State Key Laboratory of LIESMARS, Wuhan University. His current research interests include low-level computer vision and computational photography.
\end{IEEEbiographynophoto}

\begin{IEEEbiographynophoto}{\textbf{Gui-Song Xia}}
	(guisong.xia@whu.edu.cn) received his Ph.D. degree in image processing and computer vision from CNRS LTCI, T{\'e}l{\'e}com ParisTech, Paris, France, in 2011. From 2011 to 2012, he has been a Post-Doctoral Researcher with the Centre de Recherche en Math{\'e}matiques de la Decision, CNRS, Paris-Dauphine University, Paris, for one and a half years. He is currently working as a full professor in computer vision and photogrammetry at Wuhan University. He has also been working as Visiting Scholar at DMA, {\'E}cole Normale Sup{\'e}rieure (ENS-Paris) for two months in 2018. His current research interests include mathematical modeling of images and videos, structure from motion, perceptual grouping, and remote sensing imaging. He serves on the Editorial Boards of the journals Pattern Recognition, Signal Processing: Image Communications, and EURASIP Journal on Image \& Video Processing.	
\end{IEEEbiographynophoto}

\begin{IEEEbiographynophoto}{\textbf{Liangpei Zhang}}
	(zlp62@whu.edu.cn) received his B.S. degree in physics from Hunan Normal University, Changsha, China, in 1982; an M.S. degree in optics from the Xi’an Institute of Optics and Precision Mechanics, Chinese Academy of Sciences, Xi’an, in 1988; and a Ph.D. degree in photogrammetry and remote sensing from Wuhan University, China, in 1998. He is currently the head of the Remote Sensing Division, State Key Laboratory of Information Engineering in Surveying, Mapping, and Remote Sensing, Wuhan University. He is also a Chang-Jiang Scholar chair professor appointed
	by the Ministry of Education of China and a principal scientist for the China State Key Basic Research Project (2011-2016), appointed by China’s Ministry of National Science and Technology to lead the country’s remote sensing program. He has more than 500 research papers and five books to his credit and holds 15 patents. He is currently serving as an associate editor of IEEE Transactions on Geoscience and Remote Sensing. He is a Fellow of the IEEE.
\end{IEEEbiographynophoto}

\vfill
\vfill
\vfill
\vfill
\vfill








	\bibliographystyle{IEEEtran}
	\bibliography{IEEEabrv,UAVRSbibfileCR} 

\end{document}